\documentclass{article}

\usepackage{microtype}
\usepackage{graphicx}
\usepackage{subfigure}
\usepackage{booktabs} 
\usepackage{hyperref}
\usepackage{amsthm}

\usepackage[accepted]{icml2024preprint}

\usepackage{amsmath}
\usepackage{amssymb}
\usepackage{mathtools}
\usepackage{amsthm}
\usepackage{threeparttable}
\usepackage{bm}
\usepackage[capitalize,noabbrev]{cleveref}

\theoremstyle{plain}
\newtheorem{theorem}{Theorem}[section]

\newtheorem{corollary}[theorem]{Corollary}
\theoremstyle{definition}
\newtheorem{definition}[theorem]{Definition}
\newtheorem{assumption}[theorem]{Assumption}
\theoremstyle{remark}

\usepackage[textsize=tiny]{todonotes}

\icmltitlerunning{Demonstration-Guided Multi-Objective Reinforcement Learning}

\begin{document}

\twocolumn[
\icmltitle{Demonstration-Guided Multi-Objective Reinforcement Learning}



\icmlsetsymbol{equal}{*}

\begin{icmlauthorlist}
\icmlauthor{Junlin Lu}{ug}
\icmlauthor{Patrick Mannion}{ug}
\icmlauthor{Karl Mason}{ug}
\end{icmlauthorlist}

\icmlaffiliation{ug}{Department of Computer Science, University of Galway, Galway, Ireland}

\icmlcorrespondingauthor{Junlin Lu}{J.Lu5@universityofgalway.ie}

\icmlkeywords{Multi-objective Reinforcement Learning, Demonstration-Guided Reinforcement Learning, ICML}

\vskip 0.3in
]



\printAffiliationsAndNotice{}  

\begin{abstract}
Multi-objective reinforcement learning (MORL) is increasingly relevant due to its resemblance to real-world scenarios requiring trade-offs between multiple objectives. Catering to diverse user preferences, traditional reinforcement learning faces amplified challenges in MORL. To address the difficulty of training policies from scratch in MORL, we introduce demonstration-guided multi-objective reinforcement learning (DG-MORL). This novel approach utilizes prior demonstrations, aligns them with user preferences via corner weight support, and incorporates a self-evolving mechanism to refine suboptimal demonstrations. Our empirical studies demonstrate DG-MORL's superiority over existing MORL algorithms, establishing its robustness and efficacy, particularly under challenging conditions. We also provide an upper bound of the algorithm's sample complexity.
\end{abstract}

\section{Introduction}
\label{sec:Introduction}
Single-objective reinforcement learning (SORL) has exhibited promise in various tasks, such as Atari Games \cite{mnih2013playing}, robotics \cite{schulman2017proximal}, drone racing \cite{kaufmann2023champion}, and smart home energy management \cite{lu2022multi}. However, most real-world challenges are not confined to single objectives but often encompass multiple objectives. For example, there is always a need to make a trade-off between safety and time efficiency in vehicle operation or to balance heating-energy usage and cost-saving for home heating. These examples underscore the necessity to extend SORL into a multi-objective setting, known as multi-objective reinforcement learning (MORL) \cite{van2014multi,hayes2022practical}. In MORL, the target is to learn an optimal policy set that maximizes the cumulative discounted reward vector. Users can set preferences to scalarize this reward vector, aiding in the comparison of different objectives. 

However, MORL introduces additional complexities and significantly magnifies the challenges already present in SORL. The challenges include: \textbf{(i)} \textit{Sparse reward} \cite{oh2018self,ecoffet2019go,ecoffet2021first,wang2023learning}; \textbf{(ii)} \textit{Hard beginning}, where at the onset of training, policies often face substantial challenges in improving \cite{li2022reinforcement,uchendu2023jump}; \textbf{(iii)} \textit{Derailment}, where the agent may steer off the path returning to promising areas of state space due to stochasticity \cite{ecoffet2019go}. We elaborate on these challenges in Appendix \ref{subsec:SORL Challenges}.

Essentially, these issues—sparse feedback, lack of prior knowledge, and forgetfulness—represent different facets of the same underlying problem: low efficiency in exploration. A potential solution is to leverage prior demonstrations as guidance. This approach can be instrumental in providing meaningful feedback, integrating expert knowledge, and guiding the agent back toward promising areas of the state space. Unfortunately, there are currently no established methods for using prior demonstrations as guidance in MORL. Existing techniques are exclusively designed for SORL \cite{ecoffet2019go,ecoffet2021first,uchendu2023jump}. We attribute this gap to several key obstacles in applying demonstration as guidance within MORL: \textbf{(i)} \textit{Demonstration-preference misalignment}: where the preference rendering a demonstration optimal is unknown beforehand; \textbf{(ii)} \textit{Demonstration deadlock}: the paradox that the requirement to have the number of prior demonstrations \textit{n} is contingent upon the outcome of the training process itself. This stops directly extending the SORL demonstration utilization method to MORL; \textbf{(iii)} \textit{Sub-optimal demonstration}: there is a probability that sub-optimal demonstrations will lead to sub-optimal policy. These challenges are further discussed in Appendix \ref{subsec:Challeng SORL to MORL}.

In this paper, we propose a novel MORL algorithm, i.e. \textit{demonstration-guided multi-objective reinforcement learning} (DG-MORL) that utilizes demonstrations as guidance to improve exploration efficiency while overcoming those obstacles. The demonstrations can take any form as long as they can be converted to action sequences. This makes it adaptable for utilizing data from either human experts, a pre-trained policy, or even hand-crafted rule-based trajectories. We evaluate demonstrations initialized by different sources and experimentally prove they all work. To the best of our knowledge, DG-MORL is the first MORL algorithm using demonstration as guidance. It surpasses the state-of-the-art MORL algorithms in complex MORL tasks. The key contributions of our work include:

\textbf{(i) }We propose the DG-MORL algorithm which harnesses demonstrations as guidance to enhance both the effectiveness and efficiency of MORL training.

\textbf{(ii) }We introduce a self-evolving mechanism to adapt and improve the quality of guiding data by gradually transferring from prior generated demonstrations to the agent's self-generated demonstrations. This mechanism safeguards the agent against sub-optimal demonstrations and helps it learn high-performing policies.

\textbf{(iii)} Rather than requiring a large and costly demonstration set, our DG-MORL algorithm only needs a small number of demonstrations and achieves \textit{few-shot} manner in training. When there is a significant absence of prior generated demonstrations, it can still outperform the state-of-the-art MORL algorithm with a tolerable decline in performance. 

\textbf{(iv)} DG-MORL algorithm offers a universal framework that can be utilized as an enhancement to any MORL algorithm, equipping it with the capability to utilize demonstrations as a form of assistance.

We outline the related work in Section \ref{sec:Related Work}. In Section \ref{sec:Preliminaries} we present the preliminary of this paper. In Section \ref{sec:Demonstration-Guided Multi-Objective Reinforcement Learning}, we formally introduce the DG-MORL algorithm. We illustrate the experiment setting, demonstrate and discuss the experiment results in Section \ref{sec:Experiments}. In Section \ref{sec:Conclusion}, we conclude the paper. 

\section{Related Work}
\label{sec:Related Work}
\subsection{Multi-Objective Reinforcement Learning}
\label{subsec:Multi-Objective Reinforcement Learning}
MORL methods are categorized as \textit{single-policy} methods and \textit{multi-policy} methods \cite{yang2019generalized}. They are distinguished by the number of policies to solve the problem. In single-policy methods, the MORL problem is reduced to a SORL problem by using a preference to scalarize the vectorized reward feedback. The preference is supposed to be given, and therefore a standard SORL policy improvement approach can be implemented \cite{mannor2001steering, tesauro2007managing,van2013scalarized}. However, when the preference is not known beforehand, these single-policy algorithms cannot work properly. 

The multi-policy methods aim at learning a policy set to generate an approximation of the Pareto front. An intuitive and commonly used methodology is to train a single policy multiple times upon the different preferences \cite{roijers2014linear, mossalam2016multi}, while Pirotta et al. \cite{pirotta2015multi} introduced a policy-based algorithm to learn a manifold-based optimal policy. These methods suffer from the curse of dimensionality when the preference space grows large. This limits its scalability to complex tasks. To overcome this limitation, methods taking preference as input, while using a single neural network to approximate a general optimal policy for various preferences were proposed \cite{abels2019dynamic,yang2019generalized,kallstrom2019tunable}. Abels et al. and Yang et al. use vectorized value function updates. Specifically, in the work of Yang et al., the policy for a given preference can be improved using experiences from other preferences. Distinguished from the other two methods, Kallstrom et al. use a scalarized value function rather than a vectorized value function approximation. By directing the policy training with updating the gradient according to the alignment of value evaluation and preference, the general policy for MORL methodology is further improved \cite{basaklar2022pd,alegre2023sample}. Basaklar et al. used a cosine similarity between Q-values and preference weight to achieve an efficient update, while Alegre et al. used generalized policy improvement (GPI) to guide the policy to update along with the most potential direction in preference space. 

This work extends the single-network multi-policy MORL method by incorporating demonstrations to guide and enhance the efficiency of the exploration and training.

\subsection{Prior demonstration Utilization Approaches}
\label{subsec: prior demonstration Utilization Approaches}
In literature, methods using prior policies for RL initialization usually are a combination of imitation learning (IL) and RL. One of the direct ways to initialize an RL agent is to use behavior cloning to copy the behavior pattern of the prior policy \cite{rajeswaran2017learning, lu2022multi}. This method, though can effectively help with the beginning stage of RL training, cannot work well with value-based RL frameworks \cite{uchendu2023jump}. Furthermore, behavior cloning cannot work well with multi-objective settings if the preference is not known beforehand.

Previous studies \cite{vecerik2017leveraging,nair2018overcoming,kumar2020conservative} have proposed methods for integrating prior data into replay memory to help with training. These techniques have been advanced through the use of offline datasets for pre-training, followed by policy fine-tuning \cite{nair2020awac,lu2022aw}. Ball et al. proposed a simple way to use offline data to help with online RL \cite{ball2023efficient} to reduce the size of the offline dataset. However, these methods typically require an offline dataset. In contrast, our approach operates under less stringent assumptions, relying only on a few selected demonstrations.

Regrettably, the aforementioned methods have not been adapted to or validated within MORL settings. This absence of extension and empirical validation hinders direct comparisons between our method and these approaches. Additionally, the field of multi-objective imitation learning remains under-explored, presenting another limitation for comparative analysis. Therefore, our algorithm is evaluated in comparison with other existing MORL algorithms.

\subsection{Effective Exploration Approaches}
\label{subsec: Effective Exploration Approaches}
As we have introduced, the exploration efficiency may have much influence at the beginning stage of training. Several approaches have been introduced to improve the exploration efficiency, including reward shaping \cite{ng1999policy,mannion2017knowledge}, curiosity-driven exploration \cite{barto2004intrinsically}, and model-based reinforcement learning (MBRL) \cite{polydoros2017survey,alegre2023sample}. 

However, these approaches still have notable shortcomings. For example, the reliance on domain expertise for reward shaping \cite{devidze2022exploration}; The inherent vulnerability trapped in local optima in curiosity-driven exploration \cite{ecoffet2021first}; The computation overload and inaccuracy of environment model construction for MBRL. 

Using prior demonstration has the following advantages to mitigate these drawbacks. It allows for the integration of domain knowledge without an explicit reward-shaping function. This can significantly alleviate the reliance on domain expertise; With the guidance of demonstration, the agent can navigate the state space more effectively and return to promising areas that it might have otherwise struggled to revisit before the curiosity intrinsic signal is used up; Using prior demonstration does not require the construction of an environment model, as is the case in MBRL. This frees the policy from computational demands and model uncertainties associated with model-based methods.

Unfortunately, existing demonstration-utilization methods often need pre-trained Q functions \cite{nair2018overcoming,kostrikov2021offline,lu2022learning} which are not always available. Algorithms like Go-Explore \cite{ecoffet2021first} and JSRL \cite{uchendu2023jump} do not require a pre-trained Q function. However, these approaches face limitations in MORL, primarily due to their inability to address issues such as demonstration-preference misalignment and demonstration deadlock. 

\section{Preliminaries}
\label{sec:Preliminaries}
We formalise a \textit{multi-objective Markov decision process} (MOMDP) as $\mathcal{M}:=(\mathcal{S},\mathcal{A},\mathcal{T},\gamma,\mu,\bm{R})$ \cite{hayes2022practical}, where
$\mathcal{S}$ and $\mathcal{A}$ are state and action space; $\mathcal{T}:S\times A\times S \xrightarrow[]{} [0,1]$ is a probabilistic transition function; $\gamma\in[0,1)$ is a discount factor; $\mu:S_{0}\xrightarrow[]{}[0,1]$ is the initial state probability distribution; $\bm{R}:S\times A\times S \xrightarrow[]{}\mathbb{R}^{d}$ is a vectorized reward function, providing the reward signal for the multiple ($d\geq2$) objectives.

The \textit{MORL state-action value function} under policy $\pi$ is defined as:
\begin{equation}
    \label{equ:MOQ function}
    \bm{q}^{\pi}(s,a):=\mathbb{E}_{\pi}[\sum_{t=0}\gamma^{t}\bm{r}_{t}|s,a]
\end{equation}
where $\bm{q}^{\pi}(s,a)$ is a $d$-dimensional vector denoting the expected vectorized return by following policy $\pi$, $\bm{r}_{t}$ is the reward vector on \textit{t} step.

The \textit{multi-objective value vector} of policy $\pi$ with the initial state distribution $\mu$ is:
\begin{equation}
    \label{equ:equ:MOQ function}
    \bm{v}^{\pi}:=\mathbb{E}_{s_{0}\sim\mu}[\bm{q}^{\pi}(s_{0},\pi(s_{0})]
\end{equation}
$\bm{v}^{\pi}$ is the \textit{value vector} of policy $\pi$, and the $i$-th element of $\bm{v}^{\pi}$ is the returned value of the $i$-th objective by following policy $\pi$. 

The Pareto front consists of a set of nondominated value vectors. The Pareto dominance relation ($\succ_{p}$) is:
\begin{equation}
\label{equ:Pareto dominance}
    \bm{v}^{\pi}\succ_{p}\bm{v}^{\pi'}\iff(\forall i:v_{i}^{\pi}\geq v_{i}^{\pi'})\cap (\exists i:v_{i}^{\pi}>v_{i}^{\pi'})
\end{equation}
We say that $\bm{v}^{\pi}$ is nondominated when all element in $\bm{v}^{\pi}$ are not worse than any $\bm{v}^{\pi'}$ that at least one element of $\bm{v}^{\pi}$ is greater than the other $\bm{v}^{\pi'}$.
The Pareto front is defined as:
\begin{equation}
    \label{equ:Pareto front}
    \mathcal{F}:=\{\bm{v}^{\pi}|\nexists \pi'\ s.t.\ \bm{v}^{\pi'}\succ_{p}\bm{v}^{\pi}\}
\end{equation}

An optimal policy should always be one of the set of all policies whose resultant multi-objective value vector is on $\mathcal{F}$. To select an optimal policy out of the set, some criteria need to be used for further comparison. The criteria are usually given by the user and therefore termed as \textit{user-defined preference}. To involve different user-defined preferences over the objectives, a \textit{utility function} is used \cite{hayes2022practical} to map the value vector to a scalar. The utility function is sometimes referred to as \textit{scalarization function} in literature. One of the most frequently used utility functions is the linear utility function \cite{hayes2022practical}. A linear weight vector $\bm{w}$ denoting the user preference of importance over each objective is given to scalarize $\bm{v}^{\pi}$.
\begin{equation}
    u(\bm{v}^{\pi},\bm{w})=v_{\bm{w}}^{\pi}=\bm{v}^{\pi}\cdot{\bm{w}}
\end{equation}
where $\bm{w}$ is on a simplex $\mathcal{W}: \sum_{i}^{d}w_{i}=1, w_{i}\geq0$. $\mathcal{W}$ is termed as \textit{weight space}.
With the linear utility function $u$ on $\mathcal{W}$, we can define the \textit{convex coverage set} (CCS) \cite{hayes2022practical}. The CCS is a finite convex subset of the Pareto front. Any point on CCS denotes that there exists at least one optimal policy under some linear weight vector. 
Formally a CCS is defined as: 
$CCS:=\{\bm{v}^{\pi}\in\mathcal{F}|\exists\bm{w}\ s.t.\ \forall \bm{v}^{\pi'}\in\mathcal{F},\bm{v}^{\pi}\cdot\bm{w}\geq\bm{v}^{\pi'}\cdot\bm{w}\}$.
$CCS$ has all the nondominated values when a linear utility function is used. The optimal policy set is resultant values on the $CCS$.

\begin{figure}[ht]
\begin{center}
\centerline{\includegraphics[width=\columnwidth]{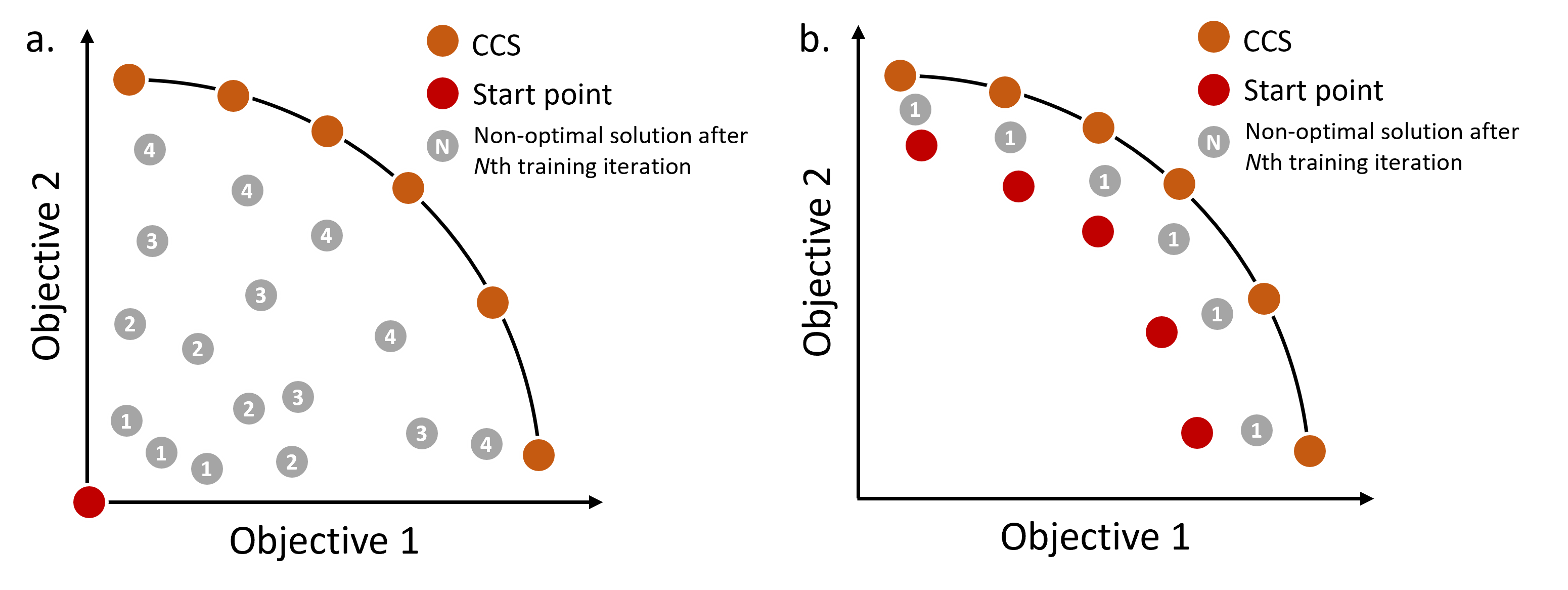}}
\caption{(a) Training process of traditional MORL; (b) Training process of demonstration-guided MORL}
\label{fig: trad MORL}
\end{center}
\vskip -0.4in
\end{figure}

In the cutting-edge MORL methods \cite{yang2019generalized, alegre2023sample}, $\pi$ is typically trained through learning a state-action vector value function, i.e. the vectorized Q function, $\bm{\mathcal{Q}}^{\pi}(s, a,\bm{w})$. This vectorized Q function can adapt to any preference weight vector. However, though these methods have achieved significant improvement upon the previous MORL algorithms, they still suffer from poor sample efficiency as they search for the $CCS$ by starting from the original point of the solution space. See Figure \ref{fig: trad MORL}(a), a traditional MORL method starts from the red point, and searches to expand to the orange CCS. 

Ideally, by giving guiding demonstrations, the training process can be transformed into a scenario depicted in Figure \ref{fig: trad MORL}(b). Here, the policy is informed by knowledge derived from the demonstrations, allowing it to commence not from the original point, but rather from the red points which represent solutions provided by the demonstrations. This shift significantly reduces the distance between the ``known" solution and the CCS, thereby decreasing sample complexity and enhancing the efficiency of the training process.

\begin{figure}[ht]
\begin{center}
\centerline{\includegraphics[width=\columnwidth]{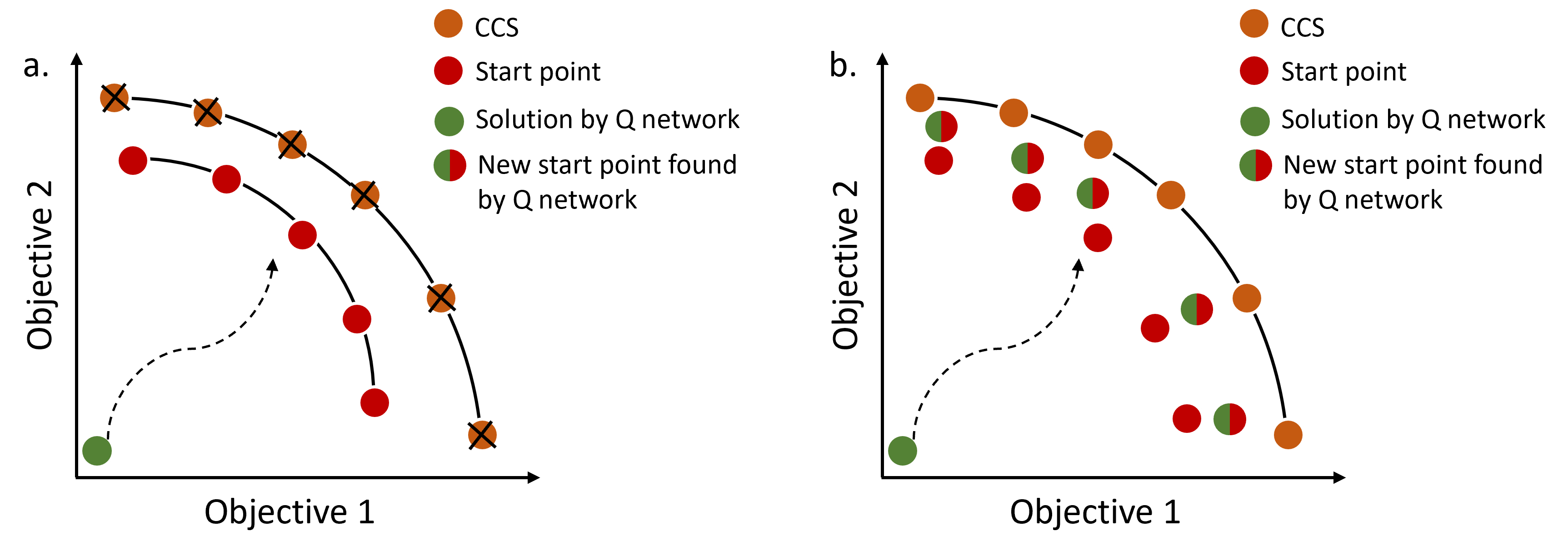}}
\caption{(a) Demonstration-guided MORL without self-evolving; (b) Demonstration-guided MORL with self-evolving}
\label{fig: Self-evolving}
\end{center}
\vskip -0.2in
\end{figure}

However, a limitation is the provided demonstrations' sub-optimality. This introduces the risk of the policy being confined within the bounds of sub-optimal demonstrations. Refer to Figure \ref{fig: Self-evolving} (a), where the green dot represents the policy approximated by a Q network. The gap between the red starting points indicating the demonstration, and the CCS can be challenging to bridge if the policy solely satisfies about meeting the performance of the prior given demonstrations. To visually represent the challenge of reaching the true CCS we have placed X marks on it.

A potential solution is that the policy may chance upon better trajectories. When the policy turns to this better guidance, it will exceed the initial demonstrations. Figure \ref{fig: Self-evolving}(b) illustrates this concept, where the dot alternating between red and green represents a demonstration identified by the Q function. Importantly, these new demonstrations are discovered by the agent itself. The initial demonstrations serve merely as a catalyst, helping the agent navigate the challenging early learning phase. This concept is formalized as the \textit{self-evolving mechanism}.

\section{Demonstration-Guided Multi-Objective Reinforcement Learning}
\label{sec:Demonstration-Guided Multi-Objective Reinforcement Learning}
In this section, we present the formal model of the DG-MORL algorithm. We begin by addressing the demonstration-preference misalignment, followed by introducing the self-evolving mechanism and then we explain how demonstration deadlock and the sub-optimal demonstration problem are addressed. Finally, we provide an in-depth description of the DG-MORL algorithm, accompanied by theoretical analysis.

\subsection{Demonstrations Corner Weight Computation}
\label{subsec:Prior Demonstrations Corner Weight Computation}
By iterating action sequences from demonstrations, the set of resultant vector returns can be calculated as the vectorized return $\mathcal{V}_{d}$. We denote the convex coverage set of the demonstrations as $CCS_{d}$.

The term \textit{corner weights} denoted as $\mathcal{W}_{corner}$ is first proposed by Roijers, D.M. \cite{roijers2016multi}. It is a finite subset of the weight simplex $\mathcal{W}$, over a set of value vectors $\mathcal{V}$.

\begin{definition} \textit{(Definition 19 of Roijers \cite{roijers2016multi})}
Given a set of value vectors $\mathcal{V}$, we can define a polyhedron $P$ by:
\begin{equation}
\label{eqn:compute corner w}
    P=\{\bm{x}\in\mathbb{R}^{d+1}|\bm{V^{+}x}\leq\bm{0},\sum_{i}w_{i}=1,w_{i}\geq0,\forall i\}
\end{equation}
where $\bm{V^{+}}$ represents a matrix in which the elements of the set $\mathcal{V}$ serve as the row vectors augmented by -1's as the column vector.
\end{definition}

The corner weights effectively partition the solution space. In informal terms, they classify the policies. The choice of the optimal policy maximizing utility for a given weight varies as we traverse across corner weights. By computing $\mathcal{W}_{corner}$ on $CCS_{d}$, we can assign corresponding weights to the demonstrations. 

Employing corner weights efficiently reduces the computational burden by calculating on a finite set $\mathcal{W}_{corner}$ rather than on an infinite weight simplex $\mathcal{W}$. Additionally, the significance of using corner weights in training is emphasized by the theorem: 

\begin{theorem} \textit{(Theorem 7 of Roijers \cite{roijers2016multi})}
\label{theo:corner w max u}
There is a corner weight $\bm{w}$ that maximizes:
\begin{equation}
\label{eqn:corner w max u}
    \Delta(\bm{w})=u^{CCS}_{\bm{w}}-\underset{{\pi\in\Pi}}{max}\ u^{\pi}_{\bm{w}}
\end{equation}
\end{theorem}

where $\Delta(\bm{w})$ is the difference between the utility of optimal policy given weight $\bm{w}$ and the utility obtained by a policy $\pi \in \Pi$, where $\Pi$ denotes the current agent's policy set. Theorem \ref{theo:corner w max u} suggests that training along a corner weight allows for faster learning of the optimal policy (set), as it offers the potential for achieving maximum improvement. Assuming the prior demonstration is the known optimal solution, by replacing the $CCS$ in equation \ref{eqn:corner w max u} with $CCS_{d}$, Equation \ref{eqn:corner w max u} can be rewritten as: $\Delta_{d}(\bm{w})=u^{CCS_{d}}_{\bm{w}}-\underset{{\pi\in\Pi}}{max}u^{\pi}_{\bm{w}}$. 
Equation \ref{eqn:select corner w} further exploits the theorem to select the most proper corner weight $\bm{w}_{c}$ from $\mathcal{W}_{corner}$ to increase the training efficiency.
\begin{equation}
\label{eqn:select corner w}
    \bm{w}_{c}=\underset{\bm{w}\in\mathcal{W}_{corner}}{argmax}\Delta(\bm{w})
\end{equation}
We have elucidated the calculation of the corner weights set and the selection of an appropriate candidate corner weight for aligning the demonstration. With this approach, the issue of \textit{demonstration-preference misalignment} is addressed.

\subsection{Self-Evolving Mechanism}
\label{subsec:Self-Evolving Mechanism}
We now propose the self-evolving mechanism. To remain consistent with the terminology used in existing literature \cite{uchendu2023jump}, the demonstrations are referred to as \textit{guide policies}, while the policies developed are called \textit{exploration policies}.

We use the notation $\Pi_{g}$ to represent the set of guide policies. Since the exploration policy set is approximated by a single neural network approximator, we denote it as $\Pi_{e}={\sum_{|\mathcal{W}|}^{i}}\pi_{e}^{\bm{w}_{i}}$. It is conditioned on the input preference weight vector and serves as the policy to be improved via learning. The agent interacts with the environment during training with a mix of guide policy and exploration policy, we denote this mixed policy as $\pi$.

The self-evolving mechanism is designed to continuously update and augment the guide policy set $\Pi_{g}$ throughout the learning process. When the agent utilizes the mixed policy $\pi$, it may discover improved demonstrations over the existing ones. This enables more effective learning from these superior demonstrations rather than persisting with sub-optimal data, thereby mitigating the impact of initially sub-optimal demonstrations. Moreover, the agent might uncover demonstrations that were not initially provided. These new demonstrations can introduce a fresh set of corner weights, enriching the guiding curriculum and addressing the \textit{demonstration deadlock}. As learning progresses, less effective demonstrations are phased out, allowing the demonstration set to evolve and improve in performance. We refer to this ongoing process of enhancement as the \textit{self-evolving} mechanism, symbolized by $\Pi_{g}\xrightarrow[]{}\Pi'_{g}$. Given that the initial demonstrations are typically sub-optimal, the guidance policy set at the end of training is likely to consist predominantly of self-generated data from the agent. The usage of the self-evolving mechanism is explained in Algorithm \ref{alg: DG-MORL Algorithm}. We also did an ablation study to show the necessity of the self-evolving mechanism, see Appendix \ref{subsec:Ablation Study}.

\subsection{Multi-Stage Curriculum}
\label{subsec:Multi-Stage Curriculum}
To make full use of the demonstration, the mixed policy $\pi$ is mostly controlled by the guide policy $\pi_{g}$ at the beginning of training. This can navigate the agent to a more promising area in the state space than randomly exploring. The control of the guide policy is diluted during training and taken over by the exploration policy. However, swapping the guide policy with the exploration policy introduces the challenge of the policy shift. A multi-stage curriculum pattern, as suggested by \cite{uchendu2023jump}, is implemented. This strategy involves gradually increasing the number of steps governed by the exploration policy, facilitating a smoother transition and mitigating the policy shift issue.

Compared with JSRL \cite{uchendu2023jump}, our approach is further extended by using the self-evolving mechanism. By eliminating outdated solutions, the MORL policy can achieve progressive improvement. Another departure is that it no longer involves swapping control between one single guide policy and the exploration policy. Instead, it entails a controlling swap between the set of guide policies to one exploration policy. Given a preference weight vector $\bm{w}$, a guide policy upon which the exploration may gain the largest improvement is selected in the next round of training as shown in Equation \ref{eqn:demonstration selection}.
\begin{equation}
\label{eqn:demonstration selection}
    \pi_{g} = \underset{\pi\in\Pi_{g}}{argmax}\Delta^{\pi}(\bm{w})=\underset{\pi\in\Pi_{g}}{argmax}(u^{\pi}_{\bm{w}}-u^{\pi_{e}}_{\bm{w}})
\end{equation}

\subsection{Algorithm}
\label{subsec: Algorithm}

We provide a detailed explanation of the DG-MORL in Algorithm \ref{alg: DG-MORL Algorithm}. To initiate the process, a set of prior demonstrations is evaluated, and the return values are added to $CCS_{d}$. Dominated values are subsequently discarded, and assume that these retained values represent ``optimal" solutions. The corner weights are then computed based on these values.

Next, a suitable guide policy is selected by using Equation \ref{eqn:demonstration selection}.
The episode horizon is denoted as $H$. The guide policy control scale is denoted as $h$, i.e. for the first $h$ steps, the guide policy controls the agent, while for the next $H-h$ steps, the exploration policy is performed. The exploration policy's performance is periodically evaluated. If it demonstrates better performance, $h$ is reduced to grant the more controllable steps to the exploration policy, and the better demonstration is popped into the demonstration repository to achieve self-evolving. The $CCS$ is periodically updated to eliminate dominated solutions. A new set of corner weights is computed based on the updated $CCS$. 

In certain instances, the guide policy may be so advanced that it becomes challenging for the exploration policy to exceed its performance. This scenario can restrict the rollback steps and therefore limit the exploration policy's touching scope of states, consequently impeding the learning efficiency. For example, the agent could find itself unable to progress beyond the initial stages of learning a solution due to the high difficulty level of the guide policy. To address this issue, we introduce the concept of a \textit{passing percentage}, denoted as $\beta\in[0,1]$. This strategy involves setting a comparatively lower performance threshold initially, making it more achievable for the agent to learn. Over time, this threshold is incrementally raised, thereby progressively enhancing the agent's performance. 
\begin{algorithm}
    \caption{DG-MORL Algorithm}
    \label{alg: DG-MORL Algorithm}
    \begin{algorithmic}[1]
    \STATE \textbf{Input:} prior demonstration $\Pi_{g}$, curriculum pass threshold $\beta$
    \STATE Evaluate the $\Pi_{g}$, add the corresponding multi-objective value vector to $CCS_{d}$
    \STATE $CCS\xleftarrow[]{}CCS_{d}$
    \WHILE{max training step not reached}
    \STATE Get corner weight set $\mathcal{W}_{corner}$=Equation \ref{eqn:compute corner w}($CCS$)
    \STATE Get candidate weight $\bm{w}_c$=Equation \ref{eqn:corner w max u}($\mathcal{W}_{corner},\Pi_{g}$)
    \STATE Sample the guide policy $\pi_{g}$=Equation \ref{eqn:demonstration selection}($\bm{w}_c,\Pi_{g}$)
    \STATE Calculate the utility threshold: $u_{\theta}=\bm{v}_{\pi_{g}}\cdot\bm{w}_{c}$
    \STATE The guide policy $\pi_{g}$ controllable scale $h=H$
        \WHILE{$h\geq0$}
            \STATE Set mix policy $\pi=\pi_{g}^{[0:h]}+\pi_{e}^{[h+1:H]}$
            \STATE Roll out $\pi$, gather the experience
            \STATE Train explore policy $\pi_{e}$
            \STATE Evaluate $\pi$: $u=\bm{v}_{\pi}\cdot\bm{w}_{c}$
            \IF{$u>u_{\theta}\cdot \beta$}
                \STATE Add the better value vector $\bm{v}_{\pi}$ to $CCS$
                \STATE $h\xleftarrow[]{}h-\Delta h$, $\Delta h$ is the rollback span
            \ENDIF
        \ENDWHILE
    \STATE Update $CCS$ and $\Pi_{g}$ by removing the dominated solutions.
    \STATE Update the curriculum pass threshold $\beta$
    \ENDWHILE
    \end{algorithmic}
\end{algorithm}

\subsection{Theoretical Analysis}
\label{subsec:Theoretical Analysis}
We provide a theoretical analysis of the algorithm sample efficiency's lower bound and upper bound.
When the exploration policy is 0-initialized (all Q-values are zero initially), a non-optimism-based method, e.g. $\epsilon$-greedy, suffers from exponential sample complexity when the horizon expands. We extend the combination lock \cite{koenig1993complexity,uchendu2023jump} under the multi-objective setting (2 objectives) and we have the following Theorem \ref{theo:Lower Bound Main}. For further explanation, please refer to Appendix \ref{subsec:Lower Bound Discussion}
\begin{theorem} (\cite{koenig1993complexity,uchendu2023jump})
\label{theo:Lower Bound Main}
   When using a 0-initialized exploration policy, there exists a MOMDP instance where the sample complexity is exponential to the horizon to see a partially optimal solution and a multiplicative exponential complexity to see an optimal solution. 
\end{theorem}
This provides the lower bound of the sample complexity for an extreme case for a MORL agent to reach a satisfying performance. We now give the general upper bound of the algorithm in Theorem \ref{theo:Upper Bound Main}.
\begin{theorem}
\label{theo:Upper Bound Main}
After $T$ rounds of training, the difference expected utility of the optimal policy $\pi^{*}$ and current policy $\pi$ is shown as:
\begin{equation}
    EU_{\pi^{*}}-EU_{\pi}\leq C(T)f(T,\mathcal{R}^{d})
\end{equation}
where $C(T)$ is the upper bound of the guide policy performance that evolves over time, $\mathcal{R}^{d}$ is the $d$-dimensionality reward space, $f(T,\mathcal{R}^{d})$ is the upper bound of the exploration algorithm performance which is determined by the total training rounds and the reward range. 
\end{theorem}
The upper bound is further discussed in Appendix \ref{subsec:Upper Bound of DG-MORL}. 

\section{Experiments}
\label{sec:Experiments}
In this section, we introduce the benchmark environments. Subsequently, we present the baseline algorithms and the detailed experiment setting. We then illustrate and discuss the experiment results.

\subsection{Benchmark Environments}
We evaluate our algorithm with commonly used challenging MORL tasks. The complexity of the tasks escalates from the instance featuring discrete state and action spaces to continuous state space and discrete action space, and finally with both continuous state and action spaces.

\textit{Deep Sea Treasure.} 
The deep sea treasure (DST) environment is commonly used in MORL setting \cite{vamplew2011empirical, abels2019dynamic, yang2019generalized,alegre2023sample}. The agent needs to trade off between the treasure value collected and the time penalty. It is featured by a discrete state space and action space. The state is the coordinate of the agent, i.e. [x,y]. The agent can move either \textit{left}, \textit{right}, \textit{up}, \textit{down}. The treasure values are aligned with literature \cite{yang2019generalized, alegre2023sample}.

\textit{Minecart.}
The Minecart environment \cite{abels2019dynamic,alegre2023sample}, is featured with continuous state space and discrete action space. The agent needs to balance among three objectives, i.e. 1) to collect ore 1; 2) to collect ore 2; 3) to save fuel. The minecart agent kicks off from the left upper corner, i.e. the base, and goes to any of the five mines consisting of two different kinds of ores and does collection, then it returns to the base to sell it and gets delayed rewards according to the volume of the two kinds of ores. 

\textit{MO-Hopper.}
The MO-Hopper, shown in Figure \ref{fig:env}(c), is an extension of the Hopper from OpenAI Gym \cite{borsa2018universal}. It is used in literature to evaluate MORL algorithm performance in robot control \cite{basaklar2022pd,alegre2023sample}. The agent controls the three torque of a single-legged robot to learn to hop. This environment involves two objectives, i.e. maximizing the forward speed and maximizing the jumping height.

In each benchmark environment, we compare the algorithms with the metric of expected utility (EU): $EU(\Pi)=\mathbb{E}_{\bm{w}\sim\mathcal{W}}[max_{\pi\in\Pi}v_{\bm{w}}^{\pi}]$. This metric is a utility-based metric \cite{zintgraf2015quality,hayes2022practical,alegre2022optimistic,alegre2023sample} used to evaluate the performance of the policy on the whole preference simplex $\mathcal{W}$\footnote{To keep aligned with the literature, we calculate EU on 100 equidistant weight vectors from $\mathcal{W}$ in each benchmark environment \cite{alegre2023sample}.}. 

For a better comparison of exploration efficiency, we impose constraints on episode lengths across benchmarks. This heightens the difficulty, compelling the agent to extract information from a shorter episode and ensuring a more rigorous assessment of candidate algorithms. See Appendix \ref{subsec:Benchmark Environment Setting} for further details of the benchmarks.

\subsection{Baseline Algorithms}
We compare our DG-MORL algorithm with two state-of-the-art MORL algorithms: generalized policy improvement linear support (GPI-LS) and generalized policy improvement prioritized Dyna (GPI-PD) \cite{alegre2023sample}. The GPI-LS and GPI-PD are based on generalized policy improvement (GPI) \cite{puterman2014markov} while the GPI-PD algorithm is model-based and GPI-LS is model-free. As indicated in the literature \cite{alegre2023sample}, the Envelope MOQ-learning algorithm \cite{yang2019generalized}, SFOLS \cite{alegre2022optimistic}, and PGMORL \cite{wurman2022outracing} are consistently outperformed by the baseline algorithms employed in our study. Therefore, if our algorithm can surpass the GPI-PD and GPI-LS baselines, it can be inferred that it also exceeds the performance of these algorithms. 

\subsection{Experiment Settings} 
 For the sake of fair comparison, the implementation https://github.com/MORL12345/DG-MORL is adapted from the existing source \cite{alegre2023sample}. Our implementation discarded all components of the original model, e.g. GPI update and Dyna, etc. except the prioritized experience replay. See details of the hyperparameters and experiment settings in Appendix \ref{subsec:Implementation Detail}. For baseline algorithms, we keep the setting as the original implementation. Please refer to the original paper for further information about those hyperparameters \cite{alegre2023sample}.

\subsection{Result}
In this section, we show the results of the EU evaluation. The result of the ablation study without the self-evolving mechanism is in Appendix \ref{subsec:Ablation Study}. The sensitivity study of different initial demonstration quantities is in Appendix \ref{subsubsec:Sensitivity to Demonstration Quantity} and the sensitivity study of qualities of the guidance policy is in Appendix \ref{subsubsec:Sensitivity to Demonstration Quality}. The impact of different rollback spans is shown in Appendix \ref{subsubsec:Impact of Different Rollback Spans}. The impact of different passing percentages of the curriculum is presented in Appendix \ref{subsubsec:Impact of Different Passing Percentages}. As the GPI-PD algorithm is a model-based version of GPI-LS, we note it as \textit{GPI-PD+GPI-LS} in the Figures to keep align with the literature \cite{alegre2023sample}. 

\subsubsection{Expected Utility Results}
\label{subsubsec:Expected Utility Results}
We randomly pick 5 seeds\footnote{Random seeds: 2, 7, 15, 42, and 78} to run the DG-MORL algorithm and the baselines to evaluate the EU. The black dashed line in each result figure shows the EU performance of the initial demonstrations. Initial demonstrations of DST, Minecart, and MO-Hopper are respectively from hard-coded action sequences, our manual play trajectories, and a pre-matured policy of a TD3 agent. The EU is evaluated with the agent solely using its exploration policy to interact with the environment. The average results are depicted with solid lines, while the range of values is illustrated by shaded areas, color-matched to the corresponding algorithms.

\begin{figure}[h!]
\begin{center}
\centerline{\includegraphics[width=\columnwidth]{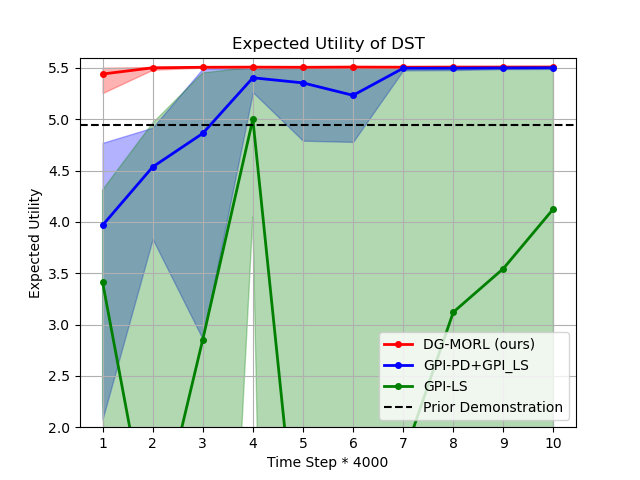}}
\caption{DST Expected Utility Result}
\label{fig:DST result}
\end{center}
\vspace{-25pt}
\end{figure}

\begin{figure}[h!]
\begin{center}
\centerline{\includegraphics[width=\columnwidth]{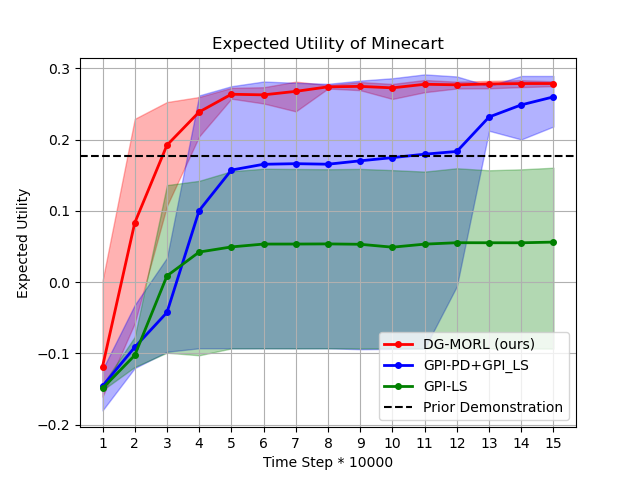}}
\caption{Minecart Expected Utility Result}
\label{fig:Minecart result}
\end{center}
\vspace{-25pt}
\end{figure}

\begin{figure}[h!]
\begin{center}
\centerline{\includegraphics[width=\columnwidth]{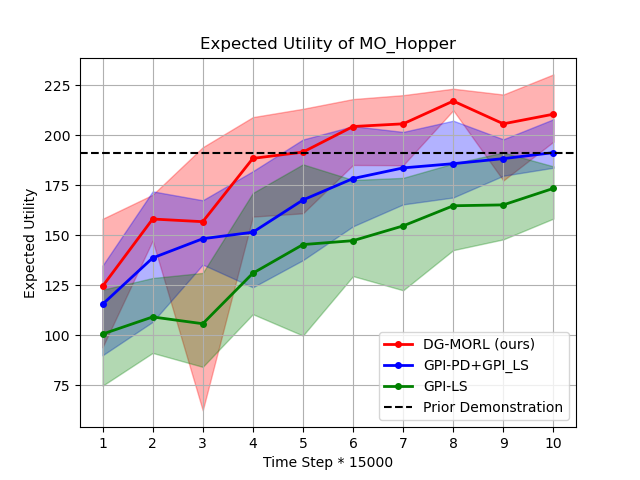}}
\caption{MO-Hopper Expected Utility Result}
\label{fig:MO-Hopper result}
\end{center}
\vspace{-25pt}
\end{figure}

\textbf{(i)} \textit{EU result for DST} 

The result of DST is shown in Figure \ref{fig:DST result}. The EU is estimated per 4000 training steps. The result shows that the DG-MORL algorithm has learned a high-performing policy in the first round of evaluation and outperforms the initial demonstration. In comparison, the GPI-PD baseline also attains a commendable level of performance in the early stages, but it is slightly inferior to DG-MORL and exhibits minor variations. The GPI-LS algorithm shows potential to reach a performance similar to GPI-PD initially, but it is markedly more vulnerable to variations, suggesting reduced robustness. Furthermore, it fails to achieve the performance levels of its counterparts. This finding aligns with expectations, given the advantage GPI-PD has in the incorporation of an environment model. In comparison, our DG-MORL outperforms all baselines regarding learning efficiency, convergence capability, and overall robustness.

\textbf{(ii)} \textit{EU result for Minecart} 

Figure \ref{fig:Minecart result} shows the result on Minecart. The EU is calculated per 10000 training steps. The results suggest that under limited exploration resources, the DG-MORL algorithm surpasses the initial demonstration performances and shows a more rapid learning curve, superior final performance, and increased robustness than baseline algorithms. Although the GPI-PD baseline eventually attains a satisfactory level of performance, it remains subpar relative to DG-MORL and suffers from instability due to varying random seed influences. Additionally, the GPI-PD does not match the learning speed of our algorithm. The GPI-LS algorithm not only faces challenges in developing an effective policy and exhibits a slower rate of learning, but also shows a vulnerability to randomness.

\textbf{(iii)} \textit{EU result for MO-Hopper} 

Figure \ref{fig:MO-Hopper result} shows the result on MO-Hopper. The EU is estimated per 15000 steps. The results reveal that DG-MORL is adept at learning the most effective policy with the fastest rate of improvement and achieves competitive results as the provided demonstration. Although it lags behind the GPI-PD baseline in the first evaluation, it quickly surpasses this with a more pronounced rate of progress, ultimately achieving the highest performance. In contrast, the GPI-PD baseline, despite showing notable improvement over time, advances at a slower pace than DG-MORL and concludes with a less effective policy. Lacking the support of an environmental model and prior demonstrations, the GPI-LS baseline gives the most inferior performance in the study.

Across all benchmark environments, the DG-MORL demonstrates superiority in terms of performance, learning efficiency, and robustness. An intriguing observation is that in environments characterized by greater continuity (both in state and action spaces), the GPI-PD and GPI-LS exhibit reduced susceptibility to variations in random seeds. This could be attributed to the fact that discrete environments are easier to be influenced by randomness. Conversely, this also underscores the robustness of our method and its insensitivity to the continuity of the environment.


\section{Conclusion}
\label{sec:Conclusion}
We proposed the pioneering DG-MORL algorithm that uses demonstrations to enhance training efficiency. We introduced the self-evolving mechanism to handle sub-optimal demonstrations. Empirical evidence confirms DG-MORL's superiority over state-of-the-art across various environments. Additionally, our experiments highlight DG-MORL's robustness by excluding self-evolving and under other unfavoured conditions. We also provided theoretical insights into the lower bound and upper bound of DG-MORL's sample efficiency. Given that real-world scenarios frequently exhibit non-linear preferences, a potential avenue for future research is addressing the limitation of DG-MORL's applicability to linear preference weights.

\section{Impact Statements}
This paper endeavors to progress multi-objective reinforcement learning by integrating demonstrations as guidance for enhanced learning performance. Our research is still in preliminary stages and based on simulated experiments. Though it will inspire various societal benefits, there are no immediate, specific social impacts necessitating emphasis at this point.


\nocite{langley00}

\bibliography{references}
\bibliographystyle{icml2023}

\newpage
\appendix
\onecolumn
\section{Explanation of Challenges}
\label{sec:Explanation of Challenges}
\subsection{SORL Challenges in MORL Setting}
\label{subsec:SORL Challenges}
\textit{Sparse reward}: In scenarios where rewards are sparse, agents struggle to acquire sufficient information for policy improvement. This challenge, widely recognized in reinforcement learning (RL) \cite{oh2018self,ecoffet2019go,ecoffet2021first,wang2023learning}, can slow down the training process, heighten the risk of reward misalignment, and may result in suboptimal policies or even training failure.

\textit{Hard beginning}: At the onset of training, policies often face substantial challenges in improving \cite{li2022reinforcement,uchendu2023jump}. This stems from the agent's lack of prior knowledge about the task. Though exploration strategy like $\epsilon$-greedy can be used to gradually mitigate this it is typically inefficient in the early stages of training.

\textit{Derailment}: The agent may steer off the path returning to promising areas of state space due to environmental stochasticity, policy uncertainty, and the randomness inherent in exploration strategies \cite{ecoffet2019go}. This derailment can impede the agent's ability to revisit these advantageous areas, thereby restricting policy improvement. 

\subsection{Challenges of Directly Extending SORL Demonstration-Utilization to MORL}
\label{subsec:Challeng SORL to MORL}
\textit{Demonstration-Preference Misalignment}: In MORL, a demonstration may be optimal under certain preferences, but detrimental under others. For an agent to effectively learn from a demonstration, it needs to know the preference that renders the demonstration optimal, as a guide of training direction. In most cases, however, even the demonstrators themselves cannot identify the specific numeric preference under which the provided demonstration is the optimal \cite{lu2023inferring}. Consequently, when an agent is provided with a set of demonstrations without associated preference information, it struggles to correctly match a demonstration with the corresponding preference factor. This can lead the agent in a misguided direction, and potentially fail the training.

\textit{Demonstration Deadlock}: In SORL, the demonstration guidance method typically relies on at least one demonstration to assist in learning the optimal policy. However, in MORL, it becomes more complex due to the presence of several, let's say \textit{n}, Pareto non-dominated policies, each optimal under different preferences. If we were to directly apply the existing demonstration-based SORL approach to MORL, it would imply the need for at least \textit{n} demonstrations to cover these various policies. The challenge is that the exact number \textit{n} remains unknown until all Pareto non-dominated policies are identified. This creates a paradox: the requirement to have the number of demonstrations \textit{n} is contingent upon the outcome of the training process itself. It can lead to a lack of necessary prior data and potentially impair the overall performance. We refer to this paradox as the \textit{demonstration deadlock}. 

\textit{Sub-optimal Demonstration}: Though demonstrations that are slightly sub-optimal can still effectively guide the training process  \cite{uchendu2023jump}, it is uncertain whether sub-optimal demonstrations can similarly work in MORL. Sub-optimal demonstrations might obstruct the training, potentially leading to the development of a sub-optimal policy, or in more severe cases, failure of the training.

\section{Experiment Setting}
\subsection{Benchmark Environment Setting}
\label{subsec:Benchmark Environment Setting}
\textit{Deep Sea Treasure.} 
\begin{figure}[ht]
\begin{center}
\centerline{\includegraphics[width=\columnwidth]{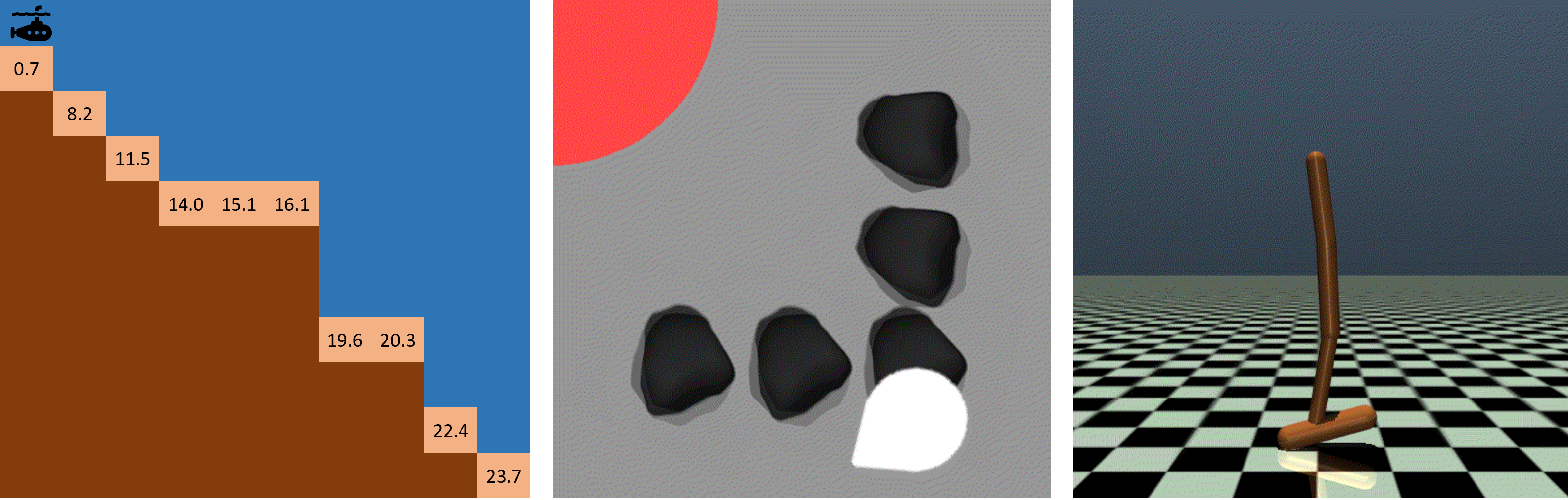}}
\caption{(a) Deep Sea Treasure; (b) Minecart; (c) MO-Hopper}
\label{fig:env}
\end{center}
\end{figure}
The deep sea treasure (DST) environment is a commonly used benchmark for the MORL setting \cite{vamplew2011empirical, abels2019dynamic, yang2019generalized,alegre2023sample}, see Figure \ref{fig:env}(a). The agent needs to make a trade-off between the treasure value collected and the time penalty. It is featured by a discrete state space and action space. The state is the coordinate of the agent, i.e. [x,y]. The agent can move either \textit{left}, \textit{right}, \textit{up}, \textit{down} in the environment. The reward vector is defined as $\bm{r}=[r_{treasure}, r_{step}]$, where $r_{treasure}$ is the value of the treasure collected, and $r_{step}=-1$ is a time penalty per step. The treasure values are adapted to keep aligned with recent literature \cite{yang2019generalized, alegre2023sample}. The discounted factor $\gamma=0.99$.

\textit{Minecart.}
The Minecart environment \cite{abels2019dynamic,alegre2023sample}, see Figure \ref{fig:env}(b), is featured with continuous state space and discrete action space. The agent needs to balance among three objectives, i.e. 1) to collect ore 1; 2) to collect ore 2; 3) to save fuel. The minecart agent kicks off from the left upper corner, i.e. the base, and goes to any of the five mines consisting of two different kinds of ores and does collection, then it returns to the base to sell it and gets delayed rewards according to the volume of the two kinds of ores. 

This environment has a 7-dimension continuous state space, including the minecart's coordinates, the speed, its angle cosine value and sine value, and the current collection of the two kinds of ores (normalized by the capacity of the minecart). The agent can select actions from the 6-dimension set, i.e. \textit{mine}, \textit{turn left}, \textit{turn right}, \textit{accelerate}, \textit{brake}, \textit{idle}. The reward signal is a 3-dimensional vector consisting of the amount of the two kinds of ore sold and the amount of fuel it has used during the episode. The reward for selling ores is sparse and delayed while the reward signal of fuel consumption is a constant penalty plus extra consumption when doing acceleration or mining.

\textit{MO-Hopper.}
The MO-Hopper, shown in Figure \ref{fig:env}(c), is an extension of the Hopper from OpenAI Gym \cite{borsa2018universal}. It is used in literature to evaluate MORL algorithm performance in robot control \cite{basaklar2022pd,alegre2023sample}. The agent controls the three torque of a single-legged robot to learn to hop. This environment involves two objectives, i.e. maximizing the forward speed and maximizing the jumping height. It has a continuous 11-dimensional state space denoting the positional status of the hopper's body components. This environment has a 3-dimensional continuous action space, i.e. the different torques applied on the three joints. We use the same reward function and discounted factor $\gamma=0.99$ \cite{alegre2023sample}. 
\begin{equation}
    r_{velocity} = v_{x}+C
\end{equation}
\begin{equation}
    r_{height} = 10(h-h_{init})+C
\end{equation}
where $v_{x}$ is the forward speed, $h$ and $h_{init}$ are the hoppers relative jumping height, $C=1-\sum_{i}a_{i}^{2}$ is the a feedback of the agent healthiness.  

To more effectively compare the exploration efficiency of our algorithm with baseline methods, we impose constraints on episode lengths across different environments. Specifically, we set the DST environment to terminate after 100 steps, the Minecart environment to conclude after 30 steps, and the MO-Hopper environment to end after 500 steps. This limitation intensifies the difficulty of the tasks, compelling the agent to extract sufficient information from a restricted number of interactions. This constraint ensures a more rigorous assessment of each algorithm's ability to efficiently navigate and learn within these more challenging conditions. The summary of benchmark settings is illustrated in Table \ref{tab:benchmarks}.

\begin{table}[t]
\caption{Benchmark Environments}
\label{tab:benchmarks}
\begin{center}
\begin{threeparttable}
\begin{tabular}{cccc}
\toprule
Environment & DST & Minecart & MO-Hopper\\
\midrule
State space $\mathcal{S}$& discrete & continuous & continuous \\
$dim(\mathcal{S})$& 2 & 7 & 11\\
Action space $\mathcal{A}$& discrete & discrete & continuous\\
$dim(\mathcal{A})$& 4 & 6 & 3\\
$dim(\mathcal{R})$\tnote{1}& 2& 3& 2\\
Episode max horizon& 100& 30& 500\\
Discounted factor\tnote{2} & 0.99 &0.98&0.99\\
Passing Percentage & 1 &1&0.8$\xrightarrow[]{}$0.98\\
Rollback Span\tnote{3}&2&2&100\\
\bottomrule
\end{tabular}
\begin{tablenotes}
    \item[1] The number of objectives / the dimension of reward vectors.
    \item[2] Consistent with baselines \cite{alegre2023sample}.
    \item[3] The number of steps rolled back when the found trajectory surpasses the given demonstration.
\end{tablenotes}
\end{threeparttable}
\end{center}
\end{table}
\subsection{Implementation Detail}
\label{subsec:Implementation Detail}
We use the same neural network architecture as the GPI-PD implementation in all benchmarks, i.e. 4 layers of 256 neurons in DST and Minecart, 2 layers with 256 neurons for both the critic and actor in MO-Hopper. We use Adam optimizer, the learning rate is $3\cdot10^{-4}$ and the batch size is 128 in all implementations. As for the exploration, we adopted the same annealing pattern $\epsilon$-greedy strategy. In DST, the $\epsilon$ anneals from 1 to 0 during the first 50000 time steps. In Minecart, the $\epsilon$ is annealed from 1 to 0.05 in the first 50000 time steps. For the TD3 algorithm doing MO-Hopper, we take a zero-mean Gaussian noise with a standard deviation of 0.02 to actions from the actor-network. All hyper-parameters are consistent with literature \cite{alegre2023sample}.

The initial demonstration data are from different sources to validate the fact that any form of demonstration can be used as it can be converted to action sequences. The demonstration for DST is from hard-coded action sequences which are sub-optimal. Minecart uses the demonstrations manually played by the author, i.e. us, while the MO-Hopper, uses the demonstration from a pre-matured policy from the TD3 algorithm.

\subsection{Baseline Algorithm}
\label{subsec:Baseline Algorithm Further}
We compare our DG-MORL algorithm with two state-of-the-art MORL algorithms: generalized policy improvement linear support (GPI-LS) and generalized policy improvement prioritized Dyna (GPI-PD) \cite{alegre2023sample}. The GPI-LS and GPI-PD are based on generalized policy improvement (GPI) \cite{puterman2014markov} while the GPI-PD algorithm is model-based and GPI-LS is model-free. As indicated in the literature \cite{alegre2023sample}, the Envelope MOQ-learning algorithm \cite{yang2019generalized}, SFOLS \cite{alegre2022optimistic}, and PGMORL \cite{wurman2022outracing}\textcolor{blue} are consistently outperformed by the baseline algorithms employed in our study. Therefore, if our algorithm can surpass these baselines, it can be inferred that it also exceeds the performance of these algorithms. 

The GPI method combines a set of policies as an assembled policy where an overall improvement is implemented \cite{barreto2017successor, barreto2020fast}. It was introduced to the MORL domain by Algre et al. \cite{alegre2022optimistic}. The GPI policy in MORL is defined as $\pi_{GPI}(s|\bm{w})\in \underset{a\in\mathcal{A}}{arg\ max}\ \underset{\pi\in\Pi}{max}q_{\bm{w}}^{\pi}(s,a)$. The GPI policy is at least as well as any policy in $\Pi$ (policy set). The GPI agent can identify the most potential preference to follow at each moment rather than randomly pick a preference as in Envelope MOQ-learning. This facilitates faster policy improvement in MORL. Furthermore, the model-based GPI-PD can find out which piece of experience is the most relevant to the particular candidate preference to achieve even faster learning. 


\section{Additional Experiments}
\label{sec:Additional Experiments}
\subsection{Ablation Study}
\label{subsec:Ablation Study}
In this section, we do an ablation study on the DG-MORL algorithm by excluding the self-evolving mechanism. The agent relies solely on prior demonstrations, as opposed to progressively enhancing the guidance demonstration with self-generated data. The ablation study uses the same random seeds set and the same demonstration set as in the EU evaluation part, see Section \ref{subsubsec:Expected Utility Results}. The results of the GPI-PD algorithm serve as a reference point, depicted as a blue dashed line. 

\begin{figure}[ht]
    \centering
    \begin{minipage}{.5\textwidth}
        \centering
        \includegraphics[width=\linewidth]{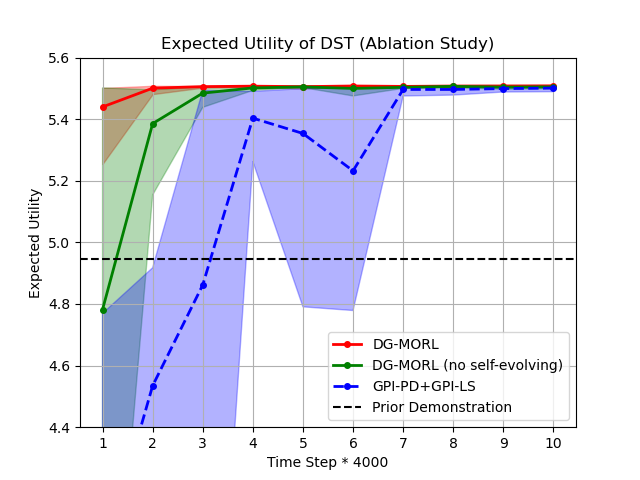}
        \caption{Ablation Study - DST}
        \label{fig:DST_ablation}
    \end{minipage}%
    \begin{minipage}{.5\textwidth}
        \centering
        \includegraphics[width=\linewidth]{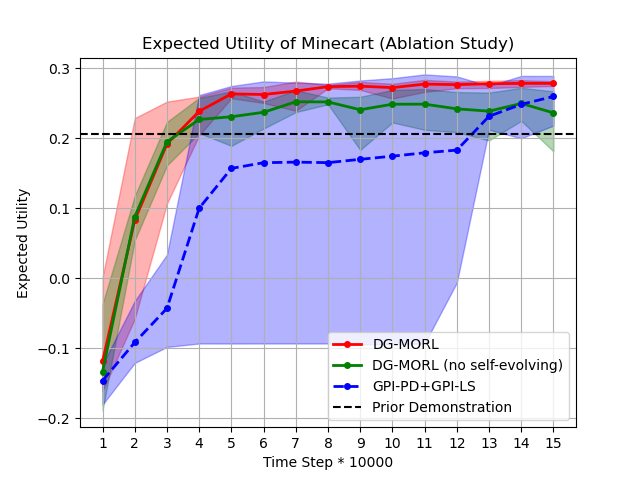}
        \caption{Ablation Study - Minecart}
        \label{fig:Minecart_ablation}
    \end{minipage}
    \begin{minipage}{.5\textwidth}
        \centering
        \includegraphics[width=\linewidth]{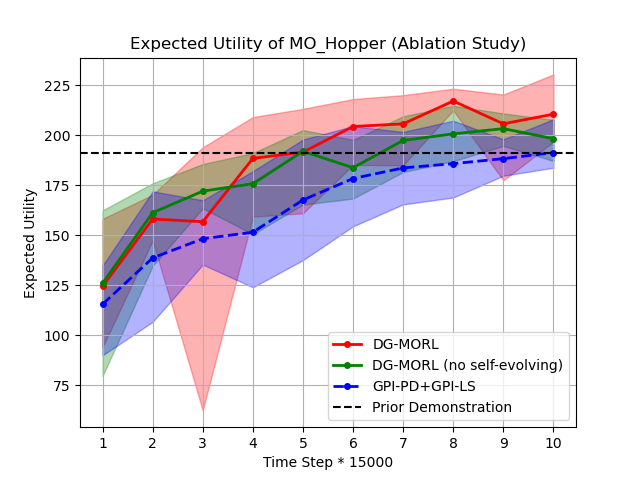}
        \caption{Ablation Study - MO Hopper}
        \label{fig:MO Hopper_ablation}
    \end{minipage}
\end{figure}

Figure \ref{fig:DST_ablation}, Figure \ref{fig:Minecart_ablation}, and Figure \ref{fig:MO Hopper_ablation} display the ablation experiment result of DG-MORL with and without the self-evolving mechanism. The absence of this mechanism makes DG-MORL fail to reach the performance levels of the fully equipped DG-MORL setup (except in DST), however, it can still overcome the bound of the prior demonstration. In the DST environment, due to the relative simplicity, the absence of the self-evolving mechanism does not impact the learning curve to reach the maximum value. However, the self-evolving mechanism guarantees a faster convergence.

In the Minecart environment, it's observed that while the DG-MORL algorithm without the self-evolving mechanism maintains high performance for the most part, its performance eventually falls below that of the GPI-PD baseline. This is indicative of a key limitation: in the absence of self-evolving, the performance of DG-MORL tends to be constrained by the quality of the initial demonstrations.

In contrast, a notable pattern emerges in the MO-Hopper environment. Initially, DG-MORL without the self-evolving mechanism manages to outperform its fully equipped counterpart. This phenomenon could be attributed to the self-evolving mechanism’s impact on the early stages of training. When this mechanism is active, it replaces the initial demonstrations with improved versions in the first few evaluation rounds. This replacement, while beneficial in the long term, makes the exploration policy’s rollback process more challenging initially, as the policy now has to surpass these enhanced demonstrations. Consequently, the rollback steps in the early stages are more difficult. However, over time, the exploration policy in the self-evolving setup catches up and eventually exceeds the performance of the non-self-evolving version, highlighting the long-term benefits of this mechanism.

Despite these nuances, it is important to note that in both scenarios: with or without the self-evolving mechanism, DG-MORL demonstrates superior performance compared to the GPI-PD baseline in terms of variations and training efficiency. The only exception is in the Minecart environment, where DG-MORL without self-evolving falls short of the GPI-PD baseline at certain points. Nevertheless, for most of the training duration, it still performs significantly better than the baseline, underlining the effectiveness of the DG-MORL algorithm in diverse settings.

The ablation study highlights the self-evolving mechanism as an integral component that significantly enhances the algorithm's performance. We provide a theory analysis in Corollary \ref{coro:self-mechnism} to show that the self-evolving mechanism can always reduce the gap between the guide policy performance and an optimal policy. Furthermore, this study substantiates the concept that utilizing demonstrations, even if they are slightly sub-optimal, can significantly benefit policy improvement, underscoring the value of this approach in the learning process.

\subsection{Sensitivity Study}
\label{subsec:Sensitivity Study}
It is widely acknowledged in machine learning that both the quantity and quality of data play crucial roles in enhancing the outcomes. Alongside these, other variables such as the rollback span and passing percentage also play significant roles in determining training performance. To understand the extent and manner in which each of these factors impacts the learning process, we have carried out a series of sensitivity studies. We continue to utilize the same three benchmark environments and the EU metric. However, the focus shifts to a comparison among DG-MORL agents rather than among other baseline algorithms. 

In the first part, we explore the sensitivity of the DG-MORL algorithm to the quantity of initial demonstrations. The second part of this section examines how the quality of data impacts the algorithm's performance. The third part of this section is about the influence of different rollback spans. The last part showcases the impact of different passing percentages \footnote{The same set of random seeds 2, 7, 15, 42, and 78 were employed. This diversity in seed selection was intended to test the robustness and generalizability of our findings across different initial conditions. In the quantity evaluation, the initial demonstrations were randomly selected from a larger set of demonstrations. To ensure consistency of the demonstration selection in this experiment, they were conducted 5 times using seed 42. This controlled approach allows for a clearer understanding of how variations in demonstration quantity affect the performance of the learning algorithm.}. 
For better comparison, the GPI-PD result is included for reference depicted by a blue dashed line. 

\subsubsection{Sensitivity to Initial Demonstration Quantity}
\label{subsubsec:Sensitivity to Demonstration Quantity}

\begin{figure}[ht]
    \centering
    \begin{minipage}{.5\textwidth}
        \centering
        \includegraphics[width=\linewidth]{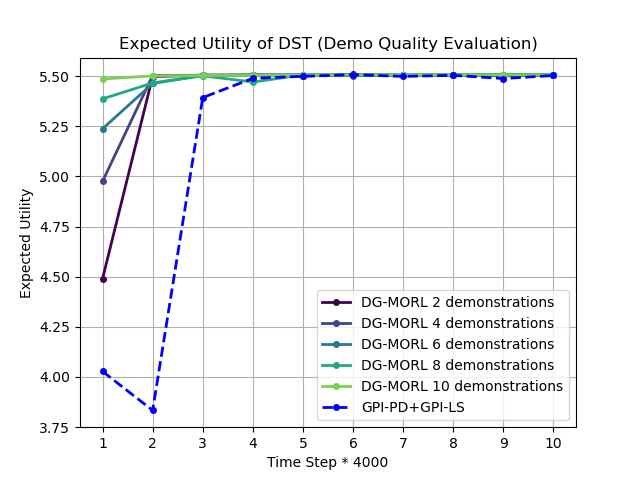}
        \caption{Demonstration Quantity Sensitivity - DST}
        \label{fig:DST_num_test}
    \end{minipage}%
    \begin{minipage}{.5\textwidth}
        \centering
        \includegraphics[width=\linewidth]{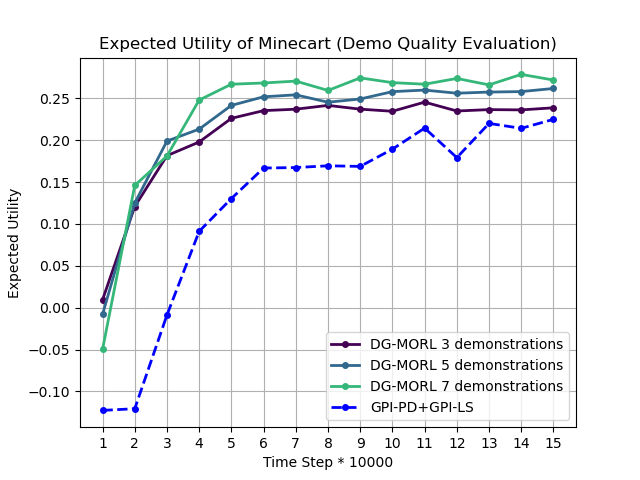}
        \caption{Demonstration Quantity Sensitivity - Minecart}
        \label{fig:Minecart_num_test}
    \end{minipage}
    \begin{minipage}{.5\textwidth}
        \centering
        \includegraphics[width=\linewidth]{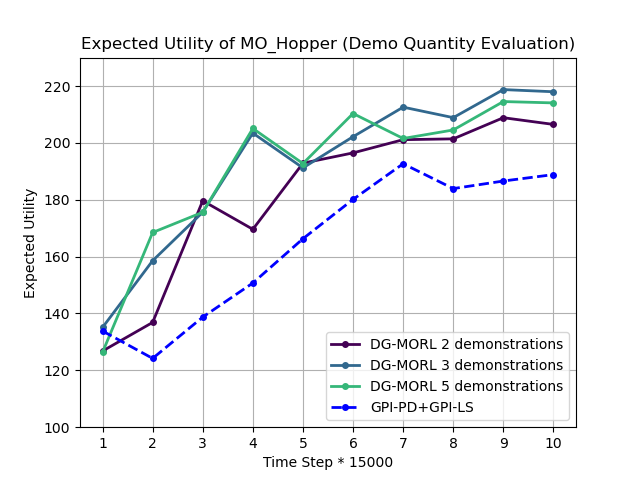}
        \caption{Demonstration Quantity Sensitivity - MO Hopper}
        \label{fig:MO_Hopper_num_test}
    \end{minipage}
\end{figure}

\begin{table}[htbp]
\caption{Demonstration Quantity Sensitivity - DST (Expected Utility)}
\label{tab:Quantity Sensitivity}
\begin{center}
\begin{tabular}{p{8pt}ccccc}
\toprule
& \multicolumn{5}{c}{Number of Prior Demonstrations}\\
\midrule
Steps &2& 4&6 & 8 & 10\\
\midrule
4k& $4.49^{+1.01}_{-1.14}$ &$4.98^{+.52}_{-1.91}$&$5.24^{+.26}_{-.87}$&$5.39^{+.11}_{-.2}$&$5.49^{+.01}_{-.01}$\\
\midrule
8k& $5.5^{+.01}_{-.02}$ & $5.5^{+.0}_{-.0}$ & $5.46^{+.04}_{-.14}$&$5.47^{+.04}_{-.14}$&$5.5^{+.01}_{-.01}$\\
\midrule
12k& $5.51^{+.0}_{-.0}$ & $5.5^{+.0}_{-.01}$ &$5.5^{+.0}_{-.0}$ & $5.5^{+.0}_{-.01}$ & $5.5^{+.0}_{-.0}$\\
\midrule
16k& $5.5^{+.0}_{-.01}$ & $5.51^{+.0}_{-.0}$ &$5.5^{+.0}_{-.01}$ & $5.47^{+.04}_{-.14}$ & $5.5^{+.0}_{-.01}$ \\
\midrule
20k& $5.5^{+.01}_{-.02}$& $5.51^{+.0}_{-.0}$ &$5.5^{+.0}_{-.01}$ & $5.51^{+.0}_{-.0}$ & $5.5^{+.0}_{-.0}$ \\
\midrule
24k& $5.51^{+.0}_{-.0}$& $5.5^{+.01}_{-.01}$ &$5.51^{+.0}_{-.01}$ & $5.51^{+.0}_{-.01}$ & $5.51^{+.0}_{-.0}$ \\
\midrule
28k& $5.51^{+.0}_{-.0}$& $5.51^{+.0}_{-.0}$ &$5.51^{+.0}_{-.0}$ & $5.5^{+.01}_{-.01}$ & $5.51^{+.0}_{-.0}$ \\
\midrule
32k& $5.51^{+.0}_{-.01}$& $5.51^{+.0}_{-.0}$ &$5.5^{+.0}_{-.0}$ & $5.51^{+.0}_{-.01}$ & $5.51^{+.0}_{-.0}$ \\
\midrule
36k& $5.5^{+.01}_{-.01}$& $5.51^{+.0}_{-.0}$ &$5.51^{+.0}_{-.0}$ & $5.51^{+.0}_{-.0}$ & $5.51^{+.0}_{-.0}$ \\
\midrule
40k& $5.51^{+.0}_{-.0}$& $5.51^{+.0}_{-.0}$ &$5.51^{+.0}_{-.0}$ & $5.51^{+.0}_{-.01}$ & $5.51^{+.0}_{-.0}$ \\
\bottomrule
\end{tabular}
\end{center}
\end{table}

\begin{table}[htbp]
\caption{Demonstration Quantity Sensitivity - Minecart (Expected Utility)}
\label{tab:Quantity Sensitivity Minecart}
\begin{center}
\begin{tabular}{cccc}
\toprule
& \multicolumn{3}{c}{Number of Prior Demonstrations}\\
\midrule
Steps &3& 5&7\\
\midrule
10k& $.01^{+.05}_{-.06}$ &$-.1^{+.06}_{-.08}$&$-.05^{+.08}_{-.1}$\\
\midrule
20k& $.12^{+.08}_{-.1}$ & $.13^{+.04}_{-.08}$ & $.15^{+.07}_{-.07}$\\
\midrule
30k& $.17^{+.08}_{-.04}$ & $.2^{+.06}_{-.08}$ &$.18^{+.07}_{-.09}$ \\
\midrule
40k& $.19^{+.06}_{-.06}$ & $.2^{+.06}_{-.05}$ &$.25^{+.01}_{-.02}$  \\
\midrule
50k& $.22^{+.06}_{-.09}$& $.23^{+.04}_{-.04}$ &$.27^{+.01}_{-.01}$ \\
\midrule
60k& $.23^{+.05}_{-.1}$& $.24^{+.03}_{-.04}$ &$.27^{+.01}_{-.01}$ \\
\midrule
70k& $.23^{+.05}_{-.11}$& $.24^{+.04}_{-.06}$ &$.26^{+.02}_{-.04}$ \\
\midrule
80k& $.24^{+.05}_{-.11}$& $.24^{+.05}_{-.04}$ &$.27^{+.01}_{-.0}$ \\
\midrule
90k& $.23^{+.05}_{-.12}$& $.22^{+.06}_{-.11}$ &$.27^{+.01}_{-.02}$ \\
\midrule
100k& $.24^{+.04}_{-.13}$& $.24^{+.04}_{-.08}$ &$.27^{+.01}_{-.02}$ \\
\midrule
110k& $.25^{+.04}_{-.11}$& $.25^{+.04}_{-.06}$ &$.27^{+.02}_{-.01}$ \\
\midrule
120k& $.24^{+.04}_{-.13}$& $.25^{+.03}_{-.05}$ &$.27^{+.0}_{-.01}$ \\
\midrule
130k& $.24^{+.04}_{-.12}$& $.25^{+.03}_{-.05}$ &$.27^{+.00}_{-.01}$ \\
\midrule
140k& $.24^{+.05}_{-.14}$& $.25^{+.03}_{-.04}$ &$.28^{+.0}_{-.01}$ \\
\midrule
150k& $.24^{+.05}_{-.14}$& $.25^{+.03}_{-.05}$ &$.27^{+.01}_{-.02}$ \\
\bottomrule
\end{tabular}
\end{center}
\end{table}

\begin{table}[htbp]
\caption{Demonstration Quantity Sensitivity - MO Hopper (Expected Utility)}
\label{tab:Quantity Sensitivity MO Hopper}
\begin{center}
\begin{tabular}{cccc}
\toprule
& \multicolumn{3}{c}{Number of Prior Demonstrations}\\
\midrule
Steps &2& 3& 5\\
\midrule
15k& $126.79^{+20.03}_{-15.23}$ &$135.1^{+15.11}_{-21.31}$&$126.34^{+26.99}_{-32.27}$\\
\midrule
30k& $136.86^{+34.41}_{-21.63}$ & $158.57^{+35.41}_{-62.43}$ & $168.48^{+23.95}_{-21.18}$\\
\midrule
45k& $179.64^{+7.7}_{-6.99}$ & $175.63^{+35.48}_{-39.68}$ &$175.61^{+24.64}_{-27.38}$ \\
\midrule
60k& $169.61^{+40.03}_{-33.78}$ & $203.45^{+17.18}_{-14.05}$ &$205.08^{+4.63}_{-3.91}$  \\
\midrule
75k& $192.86^{+18.18}_{-22.52}$& $191.21^{+26.01}_{-33.81}$ &$192.74^{+17.69}_{-11.99}$ \\
\midrule
90k& $196.48^{+12.28}_{-8.96}$& $202.18^{+23.98}_{-20.29}$ &$210.34^{+6.48}_{-5.67}$ \\
\midrule
105k& $201.13^{+18.16}_{-32.43}$& $212.57^{+12.83}_{-10.7}$ &$201.55^{+19.81}_{-41.08}$ \\
\midrule
120k& $201.42^{+9.26}_{-9.35}$& $208.88^{+16.85}_{-29.94}$ &$204.54^{+14.49}_{-24.94}$ \\
\midrule
135k& $208.86^{+5.24}_{-6.97}$& $218.75^{+11.87}_{-19.36}$ &$214.54^{+9.9}_{-9.61}$ \\
\midrule
150k& $206.5^{+9.81}_{-8.5}$& $217.99^{+13.02}_{-12.98}$ &$214.07^{+6.28}_{-10.74}$ \\
\bottomrule
\end{tabular}
\end{center}
\end{table}

Figure \ref{fig:DST_num_test}, Figure \ref{fig:Minecart_num_test} and Figure \ref{fig:MO_Hopper_num_test} shows DG-MORL algorithm's sensitivity to the quantity of initial demonstrations in the three benchmark environments separately. The figures presented depict the average results of these 5 runs for the sake of clarity.
It should be noted that scenarios without any initial demonstrations are not taken into account, as the presence of such demonstrations is a fundamental prerequisite for DG-MORL. We ensure that the process commences with a minimum number of demonstrations equivalent to the number of objectives, e.g. for DST, as there are 2 objectives, the least amount of demonstration provided should be 2 as well. More detailed evaluation results is shown in Table \ref{tab:Quantity Sensitivity} for DST, Table \ref{tab:Quantity Sensitivity Minecart} for Minecart, and Table \ref{tab:Quantity Sensitivity MO Hopper} for MO-Hopper.

Though over time, there are more demonstrations added by the self-evolving mechanism, the outcomes of our research reveal a clear relationship between the quantity of initial demonstrations and the performance of the DG-MORL algorithm, particularly evident in the DST environment. Specifically, there is a positive correlation observed: as the number of initial demonstrations increases, there is a notable enhancement in training efficiency and a corresponding improvement in policy performance.

This trend is consistent in the Minecart environment as well. Here too, the quantity of initial demonstrations plays a significant role in influencing the performance of the DG-MORL algorithm. The increased number of initial demonstrations provides more comprehensive guidance and information, which in turn facilitates more effective learning and decision-making by the algorithm. This consistency across different environments underscores the importance of initial demonstration quantity as a key factor in the effectiveness of the DG-MORL algorithm.

An intriguing observation from the Minecart experiment was that an increase in the number of initial demonstrations actually led to lower performance at the beginning of training. This phenomenon was not evident in the DST experiment, suggesting a unique interaction in the Minecart environment. We hypothesize that this initial decrease in performance is attributable to the complexity of the Minecart environment when paired with a large set of demonstrations. In such a scenario, the agent may struggle to simultaneously focus on multiple tasks or objectives presented by these demonstrations, leading to a temporary dip in performance at the early stages of training. This is indicative of an initial overload or confusion state for the agent, as it tries to assimilate and prioritize the extensive information provided by the numerous demonstrations. However, as training progresses and the agent becomes more familiar with the environment and its various tasks, it starts to overcome this initial challenge. With time, the agent effectively integrates the insights from the demonstrations, leading to improved performance. This dynamic suggests that while a wealth of demonstrations can initially seem overwhelming in complex environments, they eventually contribute to the agent's deeper understanding and better performance, underscoring the importance of considering both the quantity and nature of demonstrations in relation to the environment's complexity.

The results from the MO-Hopper environment further support the notion that a greater number of initial demonstrations tends to yield better performance. Specifically, configurations with 3 and 5 initial demonstrations outperformed the scenario with only 2 initial demonstrations. This suggests that reaching a certain threshold in the number of initial demonstrations can enhance the agent's performance.

However, an interesting deviation was observed where the performance with 3 initial demonstrations was marginally better than with 5. This outcome implies that, beyond a certain point, additional demonstrations might not always contribute positively to the learning process. Potential reasons for this include: 

Conflicting Action Selection: More demonstrations could introduce complexity in terms of conflicting action choices in similar states under different preferences. This conflict might hinder the agent's ability to learn a coherent and effective policy. 

Complexity and Long Horizon: The intrinsic complexity of the MO-Hopper environment, along with its extended horizon, might complicate the learning process. When the agent learns an effective policy that surpasses some of the guide policies, it might inadvertently ``forget" how to surpass others, particularly those aligned with different preferences.

These observations indicate that while having a sufficient number of demonstrations is beneficial, there is a nuanced balance to be struck. Too many demonstrations, especially in complex environments, can introduce new challenges that potentially offset the benefits of additional information.

Remarkably, the DG-MORL algorithm demonstrates its robust learning capabilities by outperforming the GPI-PD baseline, even when provided with a limited number of initial demonstrations. This aspect of DG-MORL underscores its efficacy in leveraging available resources to achieve superior learning outcomes, highlighting its potential for applications where abundant demonstration data may not be readily available.

\subsubsection{Sensitivity to Initial Demonstration Quality}
\label{subsubsec:Sensitivity to Demonstration Quality}

The quality of initial demonstrations emerges as another factor influencing the training process. While the self-evolving mechanism can to some degree mitigate this impact, it is still presumed that initial demonstrations of higher quality are associated with more effective learning outcomes because they may lead the agent to more potential areas in the state space.

\begin{figure}[ht]
    \centering
    \begin{minipage}{.5\textwidth}
        \centering
        \includegraphics[width=\linewidth]{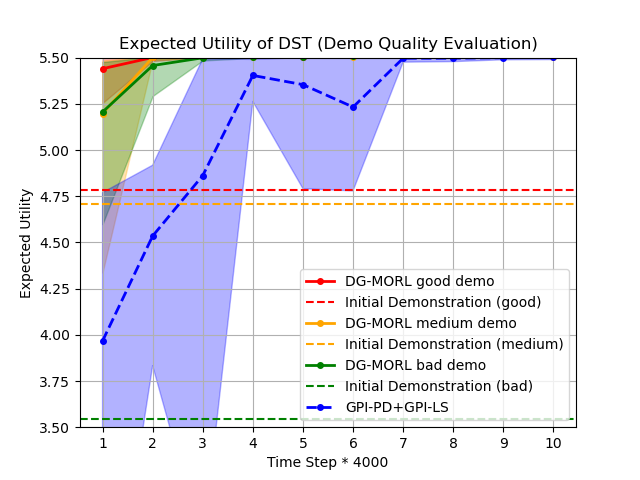}
        \caption{Demonstration Quality Sensitivity - DST}
        \label{fig:DST quality test}
    \end{minipage}%
    \begin{minipage}{.5\textwidth}
        \centering
        \includegraphics[width=\linewidth]{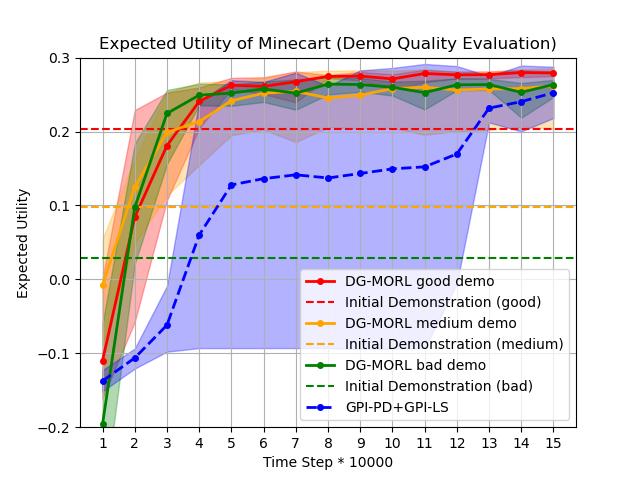}
        \caption{Demonstration Quality Sensitivity - Minecart}
        \label{fig:Minecart quality test}
    \end{minipage}
    \begin{minipage}{.5\textwidth}
        \centering
        \includegraphics[width=\linewidth]{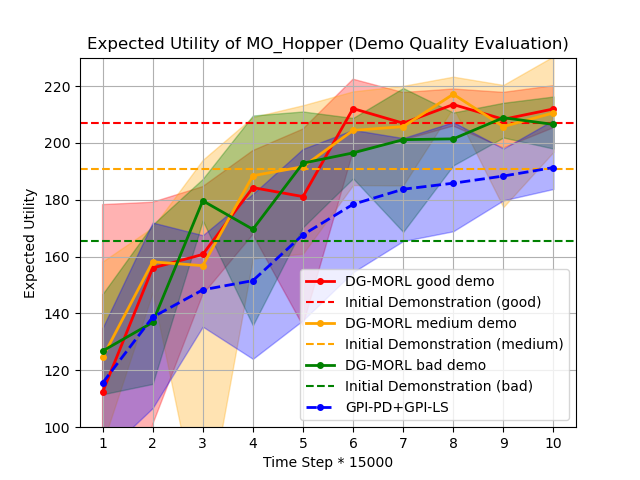}
        \caption{Demonstration Quality Sensitivity - MO Hopper}
        \label{fig:MO_Hopper quality test}
    \end{minipage}
\end{figure}

Figures \ref{fig:DST quality test}, \ref{fig:Minecart quality test}, and \ref{fig:MO_Hopper quality test} collectively illustrate the DG-MORL algorithm's responsiveness to the quality of initial demonstrations across three different environments. These figures compare the overall performance of the DG-MORL agent when initialized with varying levels of demonstration quality. Additionally, they present the EU for each set of demonstrations, indicated by dashed lines in colors corresponding to each demonstration standard.

In the DST environment, for instance, the learning outcomes for all three categories of initial demonstrations (high, medium, and low quality) eventually reach the environment's upper-performance limit. However, noteworthy differences are evident in the initial phases of training. The DG-MORL agent initialized with the highest quality demonstrations achieves significantly better performance right from the first evaluation round. The performance curve of the agent with medium-quality demonstrations shows a slightly quicker ascent compared to the one with lower-quality demonstrations. This pattern underscores the impact of initial demonstration quality on both the speed and efficiency of learning, especially in the early stages of training.

In the Minecart and MO-Hopper environments, the performance outcomes for the three sets of initial demonstrations reveal a notable observation: the quality of the demonstrations has a relatively minor influence on the final learning results. This outcome contrasts with the more pronounced effect seen in the DST environment. In Minecart and MO-Hopper, despite variations in the quality of the initial demonstrations (high, medium, and low), the differences in the final learning outcomes are less significant (but still higher). This suggests that, in these particular environments, the DG-MORL algorithm is capable of achieving comparable levels of performance regardless of the initial quality of demonstrations.

This could be attributed to the inherent characteristics of the Minecart and MO-Hopper environments, where the algorithm might have more flexibility or capability to compensate for the initial quality of demonstrations during the learning process. It highlights the adaptive nature of the DG-MORL algorithm and suggests that its performance is not solely dependent on the quality of initial demonstrations, especially in certain types of environments.

Similar to the observations made in the experiment evaluating the impact of demonstration quantity, the DG-MORL algorithm consistently outperformed the GPI-PD method in all scenarios tested for demonstration quality. This consistent superiority across different conditions and environments empirically validates the efficacy and robustness of the DG-MORL algorithm. It demonstrates that DG-MORL can effectively utilize the information provided in the demonstrations, whether of high or lower quality, to enhance its learning process and achieve superior outcomes. This finding is significant as it not only attests to the performance capabilities of DG-MORL but also reinforces its potential applicability in a wide range of real-world scenarios where the quality of available demonstrations may vary.

\subsubsection{Impact of Different Rollback Spans}
\label{subsubsec:Impact of Different Rollback Spans}

Another aspect potentially impacting training outcomes is the rollback span, which dictates the granularity of the learning process. In simpler environments, such as DST, it is hypothesized that a finer granularity leads to enhanced training performance. Conversely, in scenarios where actions are more low-level, as in the case of MO-Hopper, meaningful behavior may emerge from a sequence of actions. Here, the agent might require a larger rollback span to allow for a more extensive exploration of meaningful action combinations for appropriate behavior.

To assess this hypothesis, we conducted evaluations of the DG-MORL's performance under varying rollback spans. This approach aimed to understand better how changes in granularity affect learning efficiency and effectiveness in different environmental complexities.

\begin{figure}[ht]
    \centering
    \begin{minipage}{.5\textwidth}
        \centering
        \includegraphics[width=\linewidth]{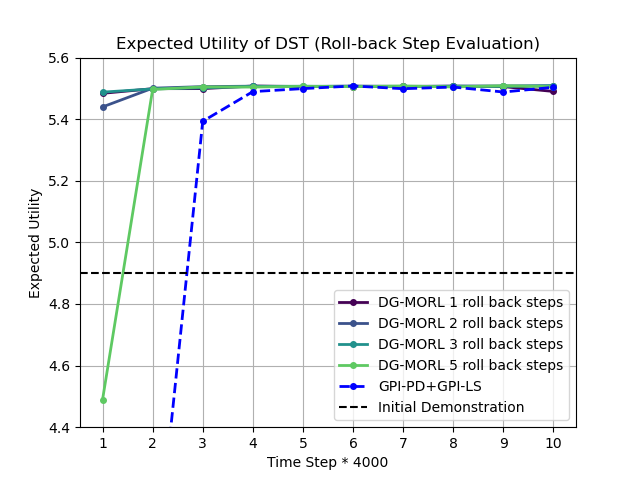}
        \caption{Rollback Span Sensitivity - DST}
        \label{fig:DST roll-back test}
    \end{minipage}%
    \begin{minipage}{.5\textwidth}
        \centering
        \includegraphics[width=\linewidth]{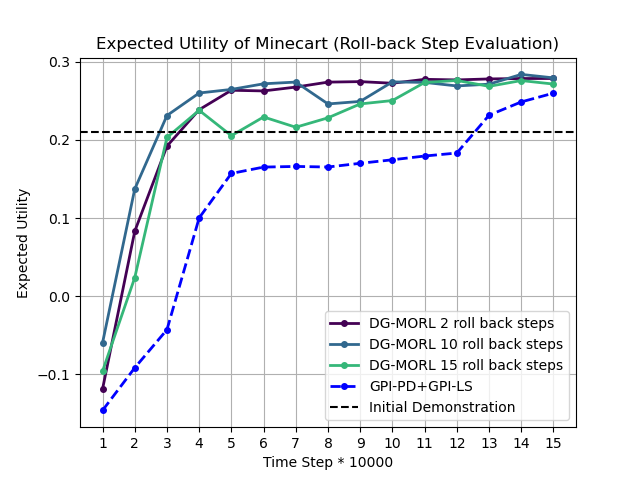}
        \caption{Rollback Span Sensitivity - Minecart}
        \label{fig:Minecart roll-back test}
    \end{minipage}
    \begin{minipage}{.5\textwidth}
        \centering
        \includegraphics[width=\linewidth]{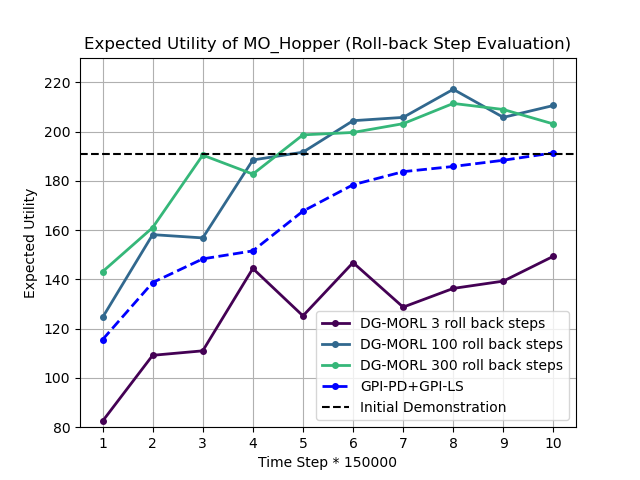}
        \caption{Rollback Span Sensitivity - MO Hopper}
        \label{fig:MO_Hopper roll-back test}
    \end{minipage}
\end{figure}
The figures presented depict the average results of these 5 runs for the sake of clarity.
Figure \ref{fig:DST roll-back test} depicts the impact of different rollback spans within the DST environment, revealing a direct correlation between finer granularity in the learning process and enhanced learning performance. Table \ref{tab:Different Rollback Span-DST} offers a detailed assessment of the effects of various rollback spans in DST. 

However, it's noteworthy that the most granular level, represented by a rollback step of 1, exhibits some performance decline in the final two evaluation rounds. This phenomenon could be attributed to potential overfitting to the demonstrations when the granularity becomes ``over-fine". Such overfitting might hinder the agent's ability to explore autonomously. This situation suggests a need for a balanced approach that effectively integrates both the guide policy and the exploration policy, indicating that an optimal level of granularity exists, which maximizes learning while avoiding the pitfalls of excessive adherence to demonstrations. 

Figure \ref{fig:Minecart roll-back test} and Table \ref{tab:Different Rollback Span-Minecart} provide insights into how different rollback spans affect training in the Minecart environment. The results roughly mirror those observed in the DST environment, particularly in how increased granularity impacts training outcomes.

When a rollback span of 10 steps is used, the agent initially shows improved performance. However, this setup leads to a noticeable decline in performance during the middle stages of training, accompanied by slightly higher variance compared to a finer granularity of rolling back for 2 steps. This suggests that while a larger rollback span can offer early benefits, it may also introduce instability and inconsistency as training progresses.

Furthermore, an even larger rollback span of 15 steps not only exhibits a similar mid-training performance drop but also lacks the initial training advantage seen with a span of 10 steps. This indicates that too large a rollback span can be detrimental right from the early stages of training, hampering the agent's ability to learn effectively. These findings highlight the importance of carefully selecting the rollback span in the Minecart environment to balance the benefits of early learning acceleration against the risks of later performance decline and increased variability.

The MO-Hopper environment is characterized by the need for the agent to consider three different torques simultaneously to achieve a cohesive behavior. Consequently, the agent's control over the hopper is at a very low level. Given this context, we hypothesize that a smaller rollback span might impede the training process. Because, if the span is too short, the agent may struggle to gain a meaningful understanding of the hopper's behavior, as its perspective is too narrowly focused on immediate, low-level torque adjustments. On the other hand, a larger rollback span could be more beneficial. It would allow the agent to observe the hopper's behavior over a broader timeframe, providing a more holistic understanding beyond mere low-level torque control. This broader perspective could enable the agent to better comprehend and learn the synthesized behavior necessary for effective control in the MO Hopper environment. 

We selected rollback spans of 3, 100, and 300 to investigate their impact on training in the MO Hopper environment. Both Figure \ref{fig:MO_Hopper roll-back test} and Table \ref{tab:Different Rollback Span-MO Hopper} provide valuable insights into this aspect. The empirical evidence gathered from these experiments supports the hypothesis that in a task requiring low-level control, such as the MO Hopper, a small rollback span is detrimental to the learning process. Specifically, when the rollback span is short (3), the agent encounters significant difficulties in learning an effective policy. This challenge is highlighted by the agent's inability to surpass the performance indicated by the initial demonstrations, represented by the black dashed line in the figure. This outcome suggests that a small rollback span limits the agent's ability to understand and integrate the broader behavioral patterns necessary for effective control in the MO Hopper environment. It emphasizes the importance of selecting an appropriate rollback span that aligns with the granularity of control and the nature of the task in reinforcement learning environments.

\begin{table}[htbp]
\caption{Impact of Different Rollback Spans - DST (Expected Utility)}
\label{tab:Different Rollback Span-DST}
\begin{center}
\begin{tabular}{p{8pt}cccc}
\toprule
& \multicolumn{4}{c}{Rollback Span}\\
\midrule
Steps &1& 2&3 & 5 \\
\midrule
4k& $5.48^{+.02}_{-.03}$ &$5.44^{+.06}_{-.19}$&$5.49^{+.01}_{-.02}$&$4.49^{+1.01}_{-.3.33}$\\
\midrule
8k& $5.5^{+.01}_{-.01}$ & $5.5^{+.01}_{-.02}$ & $5.5^{+.01}_{-.01}$&$5.5^{+.01}_{-.01}$\\
\midrule
12k& $5.5^{+.01}_{-.01}$ & $5.51^{+.0}_{-.0}$ &$5.5^{+.0}_{-.01}$ & $5.51^{+.0}_{-.0}$ \\
\midrule
16k& $5.51^{+.0}_{-.0}$ & $5.51^{+.0}_{-.0}$ &$5.51^{+.0}_{-.0}$ & $5.5^{+.04}_{-.01}$\\
\midrule
20k& $5.51^{+.0}_{-.0}$& $5.51^{+.0}_{-.01}$ &$5.5^{+.0}_{-.01}$ & $5.51^{+.0}_{-.0}$ \\
\midrule
24k& $5.51^{+.0}_{-.0}$& $5.51^{+.0}_{-.0}$ &$5.51^{+.0}_{-.0}$ & $5.51^{+.0}_{-.01}$ \\
\midrule
28k& $5.51^{+.0}_{-.01}$& $5.51^{+.0}_{-.0}$ &$5.51^{+.0}_{-.0}$ & $5.5^{+.01}_{-.0}$ \\
\midrule
32k& $5.51^{+.0}_{-.0}$& $5.51^{+.0}_{-.0}$ &$5.5^{+.0}_{-.0}$ & $5.51^{+.0}_{-.01}$ \\
\midrule
36k& $5.5^{+.01}_{-.01}$& $5.51^{+.0}_{-.0}$ &$5.51^{+.0}_{-.01}$ & $5.51^{+.0}_{-.0}$ \\
\midrule
40k& $5.49^{+.02}_{-.07}$& $5.51^{+.0}_{-.0}$ &$5.51^{+.0}_{-.0}$ & $5.51^{+.0}_{-.01}$\\
\bottomrule
\end{tabular}
\end{center}
\end{table}

\begin{table}[htbp]
\caption{Impact of Different Rollback Spans - Minecart (Expected Utility)}
\label{tab:Different Rollback Span-Minecart}
\begin{center}
\begin{tabular}{cccc}
\toprule
& \multicolumn{3}{c}{Rollback Span}\\
\midrule
Steps &2& 10&15 \\
\midrule
10k& $-.12^{+.12}_{-.04}$ &$-.06^{+.12}_{-.09}$&$-.1^{+.08}_{-.05}$\\
\midrule
20k& $.08^{+.15}_{-.14}$ &$.14^{+.03}_{-.07}$& $.02^{+.09}_{-.16}$\\
\midrule
30k& $.19^{+.06}_{-.09}$ &$.23^{+.02}_{-.04}$&$.2^{+.06}_{-.08}$\\
\midrule
40k& $.24^{+.02}_{-.03}$ &$.26^{+.01}_{-.01}$&$.24^{+.04}_{-.09}$\\
\midrule
50k& $.26^{+.01}_{-.01}$&$.26^{+.01}_{-.01}$&$.21^{+.07}_{-.27}$\\
\midrule
60k& $.26^{+.01}_{-.01}$& $.27^{+.01}_{-.01}$ &$.23^{+.04}_{-.1}$\\
\midrule
70k& $.27^{+.01}_{-.03}$& $.27^{+.01}_{-.01}$ &$.22^{+.05}_{-.13}$\\
\midrule
80k& $.27^{+.0}_{-.0}$& $.25^{+.04}_{-.13}$ &$.23^{+.05}_{-.17}$\\
\midrule
90k& $.27^{+.01}_{-.01}$& $.25^{+.04}_{-.11}$ &$.25^{+.03}_{-.07}$\\
\midrule
100k& $.27^{+.01}_{-.02}$& $.27^{+.01}_{-.01}$ &$.25^{+.02}_{-.03}$\\
\midrule
110k& $.28^{+.01}_{-.01}$& $.27^{+.01}_{-.03}$ &$.27^{+.0}_{-.0}$\\
\midrule
120k& $.28^{+.0}_{-.01}$& $.27^{+.02}_{-.02}$ &$.28^{+.01}_{-.01}$\\
\midrule
130k& $.28^{+.0}_{-.01}$& $.27^{+.02}_{-.04}$ &$.27^{+.02}_{-.01}$\\
\midrule
140k& $.28^{+.0}_{-.01}$& $.28^{+.0}_{-.0}$ &$.28^{+.01}_{-.01}$\\
\midrule
150k& $.28^{+.0}_{-.0}$& $.28^{+.01}_{-.01}$ &$.27^{+.01}_{-.02}$\\
\bottomrule
\end{tabular}
\end{center}
\end{table}

\begin{table}[htbp]
\caption{Impact of Different Rollback Spans - MO Hopper (Expected Utility)}
\label{tab:Different Rollback Span-MO Hopper}
\begin{center}
\begin{tabular}{cccc}
\toprule
& \multicolumn{3}{c}{Rollback Span}\\
\midrule
Steps &3& 100&300 \\
\midrule
15k& $82.46^{+35.46}_{-16.61}$ &$124.54^{+33.71}_{-30.47}$&$143.11^{+18.08}_{-39.13}$\\
\midrule
30k& $109.18^{+43.91}_{-31.4}$ & $158.15^{+12.12}_{-10.77}$ & $161.01^{+19.06}_{-36.91}$\\
\midrule
45k& $110.99^{+41.21}_{-22.7}$ & $156.82^{+37.21}_{-94.32}$ &$190.42^{+19.21}_{-20.35}$ \\
\midrule
60k& $144.41^{+33.55}_{-38.3}$ & $188.46^{+20.63}_{-29.09}$ &$182.72^{+30.16}_{-34.15}$ \\
\midrule
75k& $125.19^{+44.65}_{-39.65}$& $191.62^{+21.62}_{-30.6}$ &$198.64^{+10.83}_{-11.07}$ \\
\midrule
90k& $146.8^{+46.44}_{-57.1}$& $204.39^{+13.72}_{-19.22}$ &$199.62^{+10.09}_{-11.07}$ \\
\midrule
105k& $128.73^{+61.02}_{-94.84}$& $205.74^{+14.33}_{-20.69}$ &$203.18^{+9.59}_{-10.78}$ \\
\midrule
120k& $136.33^{+18.37}_{-23.65}$& $217.12^{+6.16}_{-4.63}$ &$211.4^{+11.22}_{-12.87}$\\
\midrule
135k& $139.29^{+26.12}_{-59.2}$& $205.74^{+14.63}_{-28.32}$ &$208.93^{+8.91}_{-12.7}$\\
\midrule
150k& $149.39^{+39.08}_{-32.87}$& $210.56^{+19.88}_{-13.76}$ &$203.06^{+12.47}_{-15.65}$\\
\bottomrule
\end{tabular}
\end{center}
\end{table}

\subsubsection{Impact of Different Passing Percentages}
\label{subsubsec:Impact of Different Passing Percentages}

As introduced in Section \ref{subsec: Algorithm}, the concept of a passing percentage $\beta\in[0,1]$, is introduced to alleviate the difficulty of curriculum passing. This mechanism is crucial, especially when the demonstration poses a significant challenge to be outperformed, potentially preventing the agent from progressively gaining greater control from the guiding policy and consequently hindering policy improvement. In this section, we explore how this parameter influences the learning process. Given that DST and Minecart are comparatively simpler environments, we opted for a straightforward approach by setting the passing threshold at $1$. This means the objective for the agent in these environments is to fully surpass the performance demonstrated in the initial demonstrations. Consequently, there was no need to vary the passing threshold in these simpler settings but just in the MO-Hopper environment where a progressively increasing passing percentage is utilized.

By adjusting the threshold, we can observe the balance between the need for the agent to outperform the demonstrations and the feasibility of doing so. The passing percentages, or thresholds, selected for this investigation are progressively scaled: $0.8 \rightarrow 0.98$, $0.9 \rightarrow 0.98$, and a static threshold of $1$.

\begin{figure}[htbp]
\begin{center}
\centerline{\includegraphics[width=0.8\columnwidth]{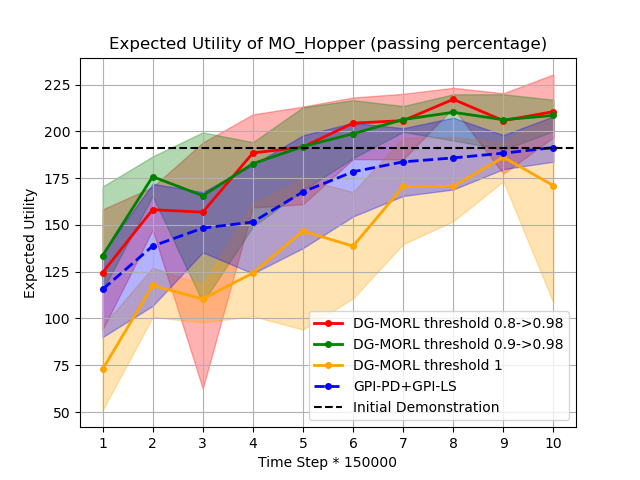}}
\caption{Curriculum Passing Percentage-MO Hopper}
\label{fig:Curriculum Passing Percentage-MO Hopper}
\end{center}
\end{figure}

The findings of our study on the effects of varying passing percentages in the MO-Hopper environment are depicted in Figure \ref{fig:Curriculum Passing Percentage-MO Hopper}. This figure illustrates that employing an incrementing passing percentage (starting from a relatively lower value) is more effective in the MO-Hopper context compared to using a static threshold value 1.

Specifically, when a static threshold of 1 was used as in DST and Minecart, the DG-MORL agent exhibited poor performance in the MO-Hopper environment. Notably, the agent was unable to surpass or even match the initial demonstration's performance. This outcome can be attributed to the high difficulty of the initial demonstration, which set an overly challenging bar for the agent. With such a high initial standard and no opportunity to progressively work towards surpassing it, the agent struggled to execute effective rollback steps and identify better demonstrations (self-evolving).

This study indicates the importance of the passing percentage as a factor in the application of DG-MORL, especially in complex environments like MO-Hopper. Adjusting the passing threshold to become incrementally more challenging allows the agent to gradually improve its performance, aligning better the complexity of the tasks at hand.
\section{Theoretical Analysis}
\label{sec:Theoretical Analysis}
\subsection{Lower Bound Discussion}
\label{subsec:Lower Bound Discussion}
\begin{figure}[ht]
\begin{center}
\centerline{\includegraphics[width=\columnwidth]{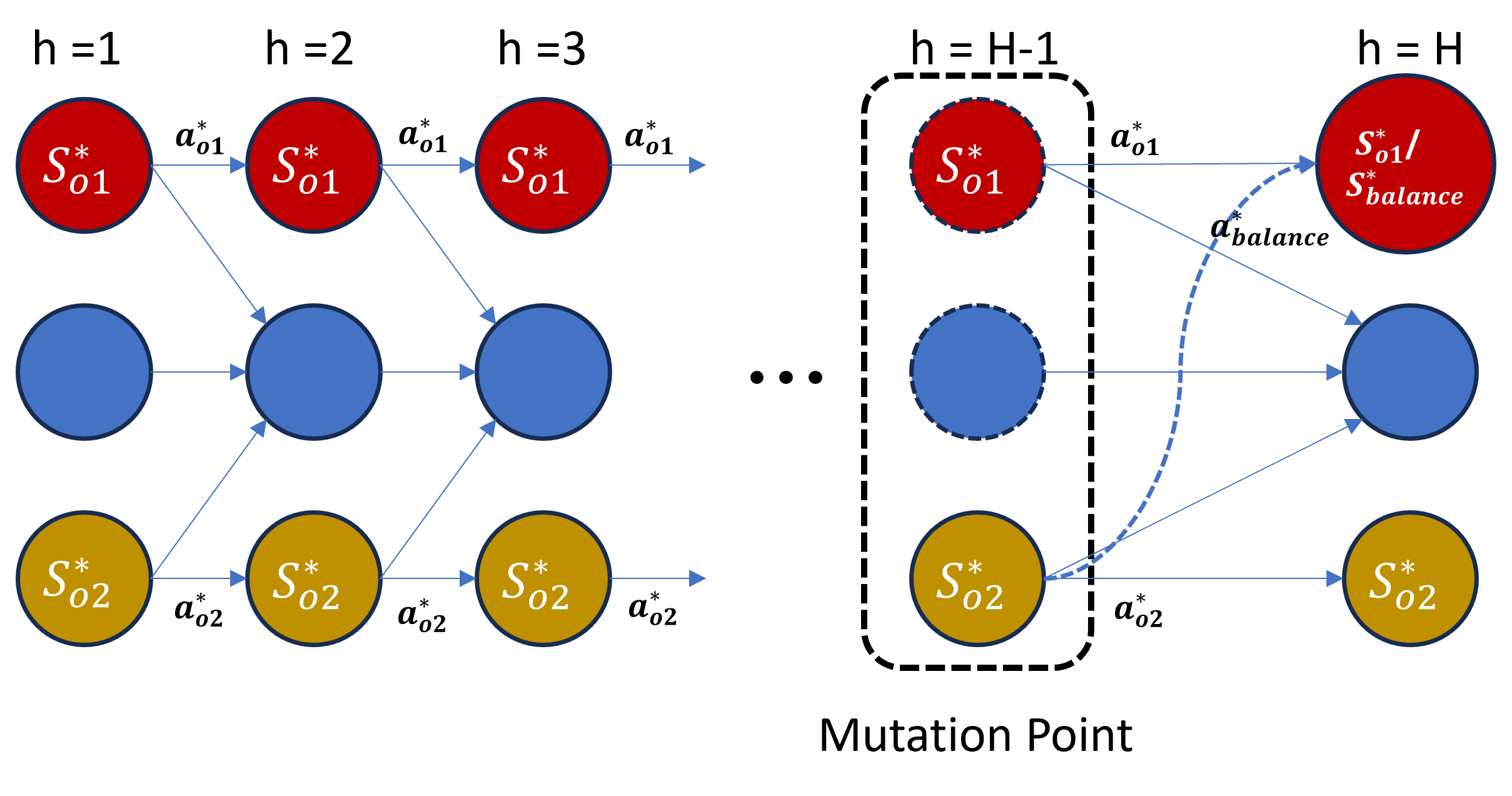}}
\caption{Lower Bound Example: MO combination lock}
\label{fig:Lower Bound Instance}
\end{center}
\end{figure}

When the exploration policy is 0-initialized (all Q-values are zero initially), a non-optimism-based method, e.g. $\epsilon$-greedy, suffers from exponential sample complexity when the horizon expands. We extend the combination lock \cite{koenig1993complexity,uchendu2023jump} under the multi-objective setting (2 objectives), and visualize an example in Figure \ref{fig:Lower Bound Instance}. 
In the multi-objective combination lock example, the agent encounters three distinct scenarios where it can earn a non-zero reward vector, each scenario is determined by a specific preference weight vector. The three preference weight vectors are $[1,0]$, $[0.5,0.5]$, and $[0,1]$. These vectors correspond to the agent's varying preferences towards two objectives: When the agent demonstrates an extreme preference for objective 1, it may receive a reward vector of $[1,0]$.
Conversely, if the agent shows an extreme preference for objective 2, it could be awarded a reward vector of $[0,1]$.

In the scenario where the agent's preference is evenly split between the two objectives, as indicated by a preference vector of $[0.5,0.5]$, there's a possibility for it to obtain a reward vector of $[0.5,0.5]$. This outcome reflects a balanced interest in both objectives. In this multi-objective combination lock scenario, the outcomes for the agent vary significantly based on its preference weights and corresponding actions. When the preference weights are set to $[1,0]$, the agent's goal is to reach the red state $s^{*}_{o{1}:h}$. To achieve this, it must consistently perform the action $a_{o1}^{*}$ from the beginning of the episode, which enables it to remain in the red state and subsequently receive a reward feedback of $[1,0]$. Similarly, with preference weights of $[0,1]$, the agent needs to execute the action $a_{o2}^{*}$ right from the start to eventually reach a reward vector of $[0,1]$. For a balanced preference of $[0.5,0.5]$, the agent is expected to take a specific action, $a_{balance}^{*}$, at step $H-1$ to transition to the state $s^{*}_{balance:h}$ and receive a balanced reward of $[0.5,0.5]$. This time point is referred to as the mutation point. If the agent derails from these prescribed trajectories, it only receives a reward of $[0,0]$. Moreover, any improper action leads the agent to a blue state, from which it cannot return to the correct trajectory (derailment). 

As the agent is 0-initialized, it needs to select an action from a uniform distribution. The preference given is also sampled from a uniform distribution. Then if the agent wants to know the optimal result for any of the three preferences, it needs at least $3^{H-1}$ samples. If it thoroughly knows the optimal solution of one of the three preferences, it can achieve an expected utility of 0.3. There exist several ($K$) paths in this MOMDP instance delivering good rewards. To find all good policies, the agent needs at least to see $K \cdot 3^{H-1}$ samples. We formalize this in Theorem \ref{theo:Lower Bound}.

The scenario where preferences precisely match the three specific weights of $[1,0]$, $[0.5,0.5]$, and $[0,1]$ is considerably rare, as preferences are typically drawn from an infinite weight simplex. In practical training situations, encountering a variety of preferences can lead to conflicting scenarios for the agent. This conflict arises because different preferences may necessitate different actions or strategies, potentially causing the agent's policy to experience ``forgetfulness" or an inability to consistently apply learned behaviors across varying preferences. Such a situation complicates the training process and increases sample complexity.

\begin{theorem} (\cite{koenig1993complexity,uchendu2023jump})
\label{theo:Lower Bound}
   When using a 0-initialized exploration policy, there exists a MOMDP instance where the sample complexity is exponential to the horizon to see a partially optimal solution and a multiplicative exponential complexity to see an optimal solution. 
\end{theorem}

\subsection{Upper Bound of DG-MORL}
\label{subsec:Upper Bound of DG-MORL}
Similar to the work of Uchendu et al. \cite{uchendu2023jump}, we start from an assumption about the guide-policy quality. 
\begin{assumption} (\cite{uchendu2023jump})
\label{assumption:Quality of guide policy}
\textit{Assume there exists a feature mapping function $\phi:\mathcal{S}\xrightarrow{}\mathbb{R}^{d}$, that for any policy $\pi$, $Q^{\pi}(s,a)$ and $\pi(s)$ depends on $s$ only through $\phi(s)$. The guide policy $\pi_{g}$\footnote{Given that the policy set $\Pi_{g}$ can be interpreted as a mixed policy tailored to specific scenarios, it retains its generality even when $\Pi_{g}$ is considered as a singular policy.} is assumed to cover the states visited by the optimal policy corresponding to the preference weight vector $w$:}
\begin{equation}
\label{eqn:quality of guide policy}
   \underset{s,h,w}{sup}\frac{d_{h,w}^{\pi^{*}}(\phi(s))}{d_{h,w}^{\pi_{g}}(\phi(s))}\leq C
\end{equation}
\end{assumption}

As we have the self-evolving mechanism, the guide policy $\pi_{g}$ also improves over time. The constant $C$ is also decreasing as the guide policy is approaching the optimal policy during training. This assumption suggests that under a preference weight vector $w$, the guide policy can see the good states which are seen by the optimal policy. 
\begin{corollary}
\label{coro:self-mechnism}
    When using the self-evolving mechanism, the upper bound of guide policy $\pi_{g}$ is no longer time-homogeneous as $\pi_{g}$ improves over time. The Equation \ref{eqn:quality of guide policy} can be rewritten as:
    \begin{equation}
        \label{time-homogeneous}
        \underset{s,h,w}{sup}\frac{d_{h,w}^{\pi^{*}}(\phi(s))}{d_{h,w}^{\pi_{g}}(\phi(s))}\leq C(t)\leq C(t_{0})
    \end{equation}
where $t$ is the training rounds, $t_{0}$ is the initial training round.
\end{corollary} 

We have provided the upper bound of the guide policy under a self-evolving mechanism. The next step is to give the upper bound of the exploration policy. When the whole horizon length is 1, the problem is reduced to a multi-objective contextual bandit (MOCB). We now establish a few key definitions to introduce the second assumption of the exploration step. One critical concept is the \textit{Pareto suboptimality gap}. This term refers to the measure of distance between the current policy's performance on a specific arm of the bandit and the true Pareto optimal solution for that arm.

\begin{definition} (Pareto suboptimality gap \cite{lu2019multi})
\label{def:Pareto suboptimality gap}
    \textit{Let $x$ be an arm in $\mathcal{X}$. The Pareto suboptimality gap $\Delta x$ is defined as the minimal scalar $\epsilon\geq0$ so that $x$ becomes Pareto optimal by adding $\epsilon$ to all entries of its expected reward.}
\end{definition}

With Definition \ref{def:Pareto suboptimality gap}, we can define the regret in a multi-objective setting, i.e. Pareto regret (PR).
\begin{definition} (Pareto regret\cite{drugan2013designing,lu2019multi})
    \label{def:Pareto regret}
    \textit{If $x_{1},x_{2},...,x_{n}$ are the arms can be pulled by the learner. The Pareto regret is: $PR(N)=\sum^{N}_{n=0}\Delta x_{n}$}
\end{definition}

We can give the performance guarantee with the help of the aforementioned definitions. Note that, we build our analysis on MOCB, which is suitable for single-step decision-making. As our algorithm works based on the Q-learning and TD3 paradigm, which is more robust in our benchmarks, i.e. sequence decision-making, it is supposed to be less likely to violate the assumption. 
\begin{theorem} (Performance guarantee for exploration algorithm in MOCB \cite{drugan2013designing, uchendu2023jump})
\label{theorem:performance guarantee for explorationoracle_CB}
\textit{In MOCB where the reward vector is constrained by the reward space $\mathcal{R}^{d}$ ($d$ is the dimensionality of the objectives)}, there exists some exploration algorithm performing a policy $\pi_{t}$ in each round $t\in[T]$, where the Pareto regret is bounded:
 \begin{equation}
        \label{eqn:Pareto regret bound}
        PR\leq f(T,\mathcal{R}^{d})
    \end{equation}
\end{theorem}

With Assumption \ref{assumption:Quality of guide policy} and Theorem \ref{theorem:performance guarantee for explorationoracle_CB}, DG-MORL can guarantee that after $T$ rounds:
\begin{equation}
    EU_{\pi^{*}}-EU_{\pi}\leq C(T)f(T,\mathcal{R}^{d})
\end{equation}
The upper bound depends on the exploration algorithm. When using the exploration algorithm as UCB \cite{drugan2013designing}, this upper bound can be written as: $EU_{\pi^{*}}-EU_{\pi}\leq \widetilde{O}(T) \times C(T)$. When using the exploration algorithm as GLM \cite{lu2019multi}, this upper bound can be written as $EU_{\pi^{*}}-EU_{\pi}\leq \widetilde{O}(d\sqrt{T}) \times C(T)$. 

\end{document}